%% file: main.tex
\documentclass[10pt,twocolumn,letterpaper]{article}

\PassOptionsToPackage{table}{xcolor}

\usepackage[datasets]{wacv}      % To produce the REVIEW version for the algorithms track
\usepackage{booktabs}
\usepackage{pifont}    % \ding ticks and crosses
\usepackage{makecell}  % multi-line header cell
\usepackage{graphicx}
\usepackage{tikz}
\usepackage{tcolorbox}
\usepackage{booktabs}
\usepackage{multirow}
\usetikzlibrary{arrows.meta}
\definecolor{arrowgray}{HTML}{555555}
\definecolor{labelgray}{HTML}{444444}
\definecolor{baseblue}{HTML}{5D8DFD}
\definecolor{mygreen}{HTML}{88B06D}
\newcommand{\cmark}{\textcolor{mygreen}{\ding{51}}}
\definecolor{myred}{HTML}{F5433D}
\newcommand{\xmark}{\textcolor{myred}{\ding{55}}}

\definecolor{softpink}{HTML}{FADADD}
\newcommand{\attr}[1]{\cellcolor{softpink}#1}
\def\revision{} %{\textcolor{blue}}

\input{preamble}

\definecolor{wacvblue}{rgb}{0.21,0.49,0.74}
\usepackage[pagebackref,breaklinks,colorlinks,allcolors=wacvblue]{hyperref}

\def\wacvPaperID{758} % *** Enter the WACV Paper ID here
\def\confName{WACV}
\def\confYear{2027}

\title{Attractive Faces Skew Text-to-Image Reward Models and Leaderboards}

\author{Harvey Mannering\\
University of Southampton\\
{\tt\small hgm1g14@soton.ac.uk}
\and
Zhiwu Huang\\
University of Southampton\\
{\tt\small zhiwu.huang@soton.ac.uk}
\and
Adam Prügel-Bennett\\
University of Southampton\\
{\tt\small apb1@soton.ac.uk}
}

\begin{document}
\maketitle
\input{sec/0_abstract}
\input{sec/1_intro}
\input{sec/2_related_work}
\input{sec/3_key_observation}
\input{sec/4_leaderboard}

\input{sec/5_reward_bias}

\input{sec/6_generation_bias}
\input{sec/7_mitigations}
\input{sec/8_dicussion}
\input{sec/9_conclusion}

\section*{Acknowledgements}
This research was mainly supported by Studentship from the Faculty of Engineering and Physical Sciences - Electronic and Computer Science at the University of Southampton. The authors acknowledge the use of the IRIDIS X High Performance Computing Facility and the Southampton-Wolfson AI Research Machine (SWARM) GPU cluster, funded by the Wolfson Foundation, together with the associated support services at the University of Southampton in the completion of this work.

{
    \small
    \bibliographystyle{ieeenat_fullname}
    \bibliography{main}
}

\appendix
\input{appendix}

\end{document}

%% file: preamble.tex
\DeclareUnicodeCharacter{221E}{\ensuremath{\infty}}

%% file: sec/0_abstract.tex
\begin{abstract}
Human preference is now central to the evaluation, guidance, finetuning, and ranking of text-to-image models. However, the susceptibility of human preferences to bias in image generation remains underexplored. We identify a key observation: for human faces, perceived attractiveness skews image quality judgments, even when comparing real photographs of comparable fidelity. In a controlled user study, annotators rank more attractive faces as higher quality, even when unattractive real faces are compared with attractive AI-generated ones. We examine the downstream consequences of this bias. Firstly, simulating Elo rankings from user study data shows attractiveness can \revision{distort} a leaderboard's ranking, elevating synthetic attractive faces over real, unattractive ones. Secondly, the same bias reappears in the human preference models. Editing 40 annotated facial attributes in isolation, we find that attractiveness is the strongest correlate of score for four of five human preference models (PickScore, MPS, HPSv2.1, HPSv3), and the strongest predictor of score in regression analysis. \revision{We further show that preference aligned generation shifts faces toward younger, more feminine faces, and can override explicit prompts for unattractive faces, thereby limiting user control.} Finally, we explore mitigation strategies, both for data collection and post training on human preference models. These findings highlight the need for preference models that disentangle image quality from human cognitive bias, enabling more reliable evaluation.
\end{abstract}

%% file: sec/1_intro.tex
\section{Introduction}
\label{sec:intro}

Text-to-image (T2I) models have recently emerged as a powerful and user-friendly class of generative models \citep{rombach2022high,ramesh2022hierarchical,saharia2022photorealistic,chen2024pixart}. T2I models can be challenging to evaluate due to the breadth of their generative capabilities, however human preference has emerged as one of the most common approaches. To capture this, T2I leaderboards such as LMArena\footnote{\url{https://arena.ai/}} present users with images from two competing models and ask them to select their preference. The resulting data is then used to establish model rankings, providing benchmarks that organizations frequently rely upon to demonstrate state-of-the-art performance. Reflecting the importance of these benchmarks, the majority of models currently on LMArena's T2I open-weight leaderboard have undergone some form of human preference alignment during post-training. Furthermore, human preference models, which are trained to predict which image a user will prefer, are now widely used for guided sampling \citep{eyring2024reno,potaptchik2026meta}, for supervised finetuning \citep{fan2023dpok,black2024training,clark2024directly}, and as evaluation metrics \citep{mannering2025noise,liao2024facescore,wang2025fairhuman,wang2025dp}. 

\begin{figure}[t]
  \centering
  \begin{minipage}[t]{0.45\columnwidth}
    \centering
    \includegraphics[width=\linewidth]{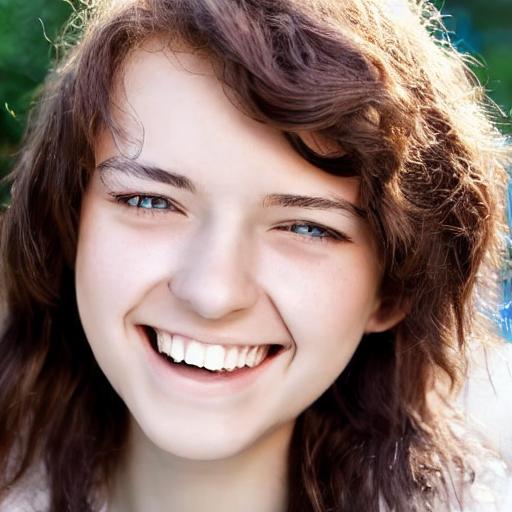}\\[5pt]
    {\sffamily\footnotesize
     \renewcommand{\arraystretch}{1.15}%
     \begin{tabular}{@{}l@{\hspace{2.5em}}r@{\hspace{0.6em}}c@{}}
       \multicolumn{3}{c}{\bfseries Stable Diffusion v1.5}\\
       \midrule
       \textcolor{labelgray}{ImageReward} & \textcolor{mygreen}{\bfseries 0.79}  & \cmark\\
       \textcolor{labelgray}{PickScore} & \textcolor{mygreen}{\bfseries 20.52}  & \cmark\\
       \textcolor{labelgray}{HPSv2.1}   & \textcolor{mygreen}{\bfseries 0.30} & \cmark\\
       \textcolor{labelgray}{MPS}       & \textcolor{mygreen}{\bfseries 9.23} & \cmark
     \end{tabular}}
  \end{minipage}\hfill
  \begin{minipage}[t]{0.45\columnwidth}
    \centering
    \includegraphics[width=\linewidth]{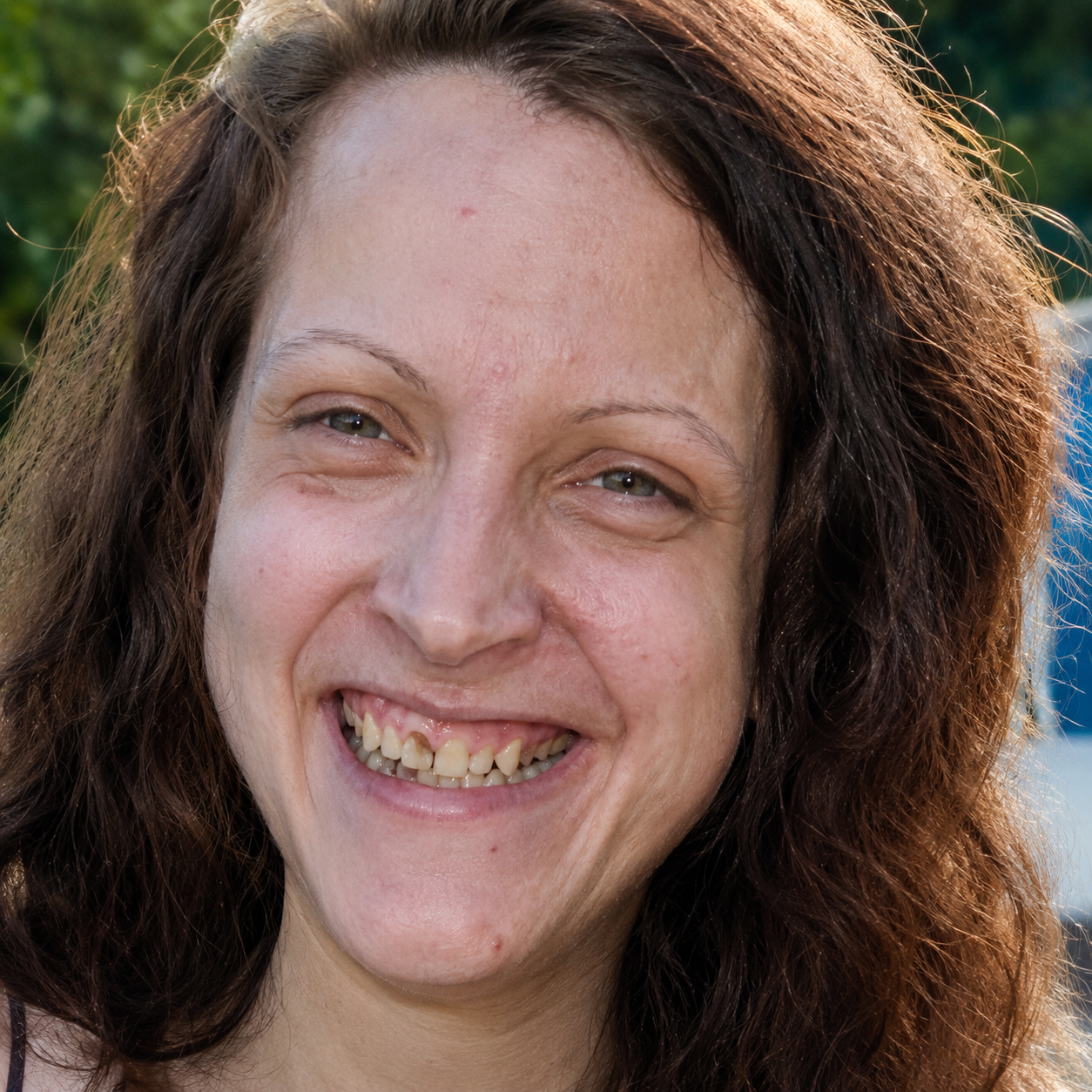}\\[5pt]
    {\sffamily\footnotesize
     \renewcommand{\arraystretch}{1.15}%
     \begin{tabular}{@{}l@{\hspace{2.5em}}r@{\hspace{0.6em}}c@{}}
       \multicolumn{3}{c}{\bfseries GPT-Image-2}\\
       \midrule
       \textcolor{labelgray}{ImageReward}   & \textcolor{myred}{\bfseries 0.51} & \xmark\\
       \textcolor{labelgray}{PickScore} & \textcolor{myred}{\bfseries 20.03}  & \xmark\\
       \textcolor{labelgray}{HPSv2.1}   & \textcolor{myred}{\bfseries 0.26} & \xmark\\
       \textcolor{labelgray}{MPS}       & \textcolor{myred}{\bfseries 7.55} & \xmark
     \end{tabular}}
  \end{minipage}
  \caption{Two AI-generated faces and the scores assigned by four widely used preference models. Stable Diffusion v1.5 (2022) was prompted for an \emph{attractive} face; \revision{GPT-Image-2 (2026) was given a blurred version of the attractive face and prompted for an \emph{unattractive} face, preserving its structure.} Preference models mark the attractive face as better despite visible artefacts.}
  \label{fig:score_comparison}
\end{figure}

In this paper, we uncover a failure mode for human preference data for face images. In a user study, we find a tendency to rank more attractive faces as better, even when both images are real and even when the more attractive face is lower fidelity. This bias appears to be robust to the instruction wording. We go on to explore the implications of this bias for the evaluation and training of image generation. Elo score simulations suggest that facial attractiveness can override image fidelity and skew rankings of AI image generators. \revision{Figure~\ref{fig:score_comparison} provides an illustrative example of the impact on human preference models, where Stable Diffusion v1.5 \citep{rombach2022high} can achieve higher scores than GPT-Image-2 \citep{openai2026chatgptimages2} simply by prompting for an ``attractive'' rather than an ``unattractive'' face, despite GPT-Image-2 being considerably newer and stronger on most metrics.} Our experiments also suggest that preference aligned generation can effectively limit user control for face generation, with the signal for an ``attractive'' face overriding the signal for following a given text prompt.

Taken together, the observed bias toward attractive faces has consequences that ripple through the entire T2I ecosystem. Beyond contaminating leaderboards and evaluation metrics, optimizing generation against these signals risks re-encoding a narrow, homogeneous beauty standard into the images AI systems produce, compounding well-documented harms linked to exposure to idealized facial imagery \citep{ramphul2018snapchat,isakowitsch2022augmented,holland2016systematic,giulio2020concept}. To the best of our knowledge, we are the first to identify, test, and quantify this bias across the human preference data, reward models, and generative outputs of T2I models.  Our contributions are:

\begin{itemize}
\item A controlled user study demonstrating that people tend to rank more attractive faces as higher quality, even when image fidelity is equal or lower.
\item We audit leaderboard systems, human preference models, and preference aligned generation, demonstrating that each have clear vulnerabilities to attractiveness bias.
\item We explore pre- and post-training mitigation through a comparative user study and evaluations across four human preference models and five T2I models.
\end{itemize}

%% file: sec/2_related_work.tex
\section{Related Work}
\label{sec:related_work}

\paragraph{Human Preference Models} Text-to-image (T2I) generation is now dominated by diffusion \citep{sohl2015deep,song2019generative,ho2020denoising} and flow matching \citep{lipman2022flow,liu2022flow} models such as Stable Diffusion \citep{rombach2022high}, Qwen Image \citep{wu2025qwen}, and FLUX \citep{labs2025flux}. To further improve these models, researchers have turned to preference alignment to better align generation with human judgments \citep{fan2023dpok,black2024training}. Human preference models act as reward functions $r_\psi(y, c)$ that predict human judgments of visual quality, aesthetic appeal, and prompt alignment for a generated image $y$ and a given caption $c$. Models such as PickScore \citep{kirstain2023pick}, ImageReward \citep{xu2023imagereward}, MPS \citep{zhang2024learning}, and the HPS series \citep{wu2023human,ma2025hpsv3} leverage large-scale human comparisons to provide their reward signals, which are now routinely used for reward-guided sampling \citep{eyring2024reno,ma2025inference,kim2025test,potaptchik2026meta}, RL and non-RL finetuning \citep{fan2023dpok,black2024training,xu2023imagereward,clark2024directly,prabhudesai2023aligning}, dataset filtering \citep{ahn2026noise}, and evaluation \citep{mannering2025noise,liao2024facescore,wang2025fairhuman,wang2025dp,wang2026diffusion}.

However, problems have started to emerge around these models. MJ-Bench \citep{chen2026mj} finds they can exhibit bias around race and gender. F-Bench is a human preference dataset for AI generated and edited faces, however, no analysis is done on the connection between preference and attractiveness and at the time of writing the dataset has not been made available \citep{liu2025f}. Most relevant to our work, \citet{guo2025position} argues that human preference models' focus on aesthetics can be problematic for user control and anti-aesthetic generation (e.g., abstract artwork and negative-valence expressions receive lower rewards) but do not identify the single most significant factor biasing scores for facial imagery. We continue this line of work in the domain of faces, finding that attractiveness is that factor, and identify its causes.

\paragraph{Attractiveness Bias}
The attractiveness halo effect refers to the well-established tendency to attribute positive characteristics, such as greater social desirability, happiness, and success, to physically attractive individuals \citep{dion1972beautiful,gulati2024beautiful}. Whether similar biases influence the perception of AI-generated images remains an open question. \citet{makowski2025too} provide an initial investigation by flashing participants with two real face images, misleadingly asking them to classify each as real or AI-generated. They find a strong linear relationship between facial attractiveness and perceived ``realness,'' suggesting that human judgments of AI-generated images may be biased toward perceiving more attractive faces as more realistic. This setting differs from preference data collection, which allows unlimited inspection time and presents images known to be AI-generated rather than asking participants to guess which are real. Our user study is designed to address this gap.

This question matters given evidence linking exposure to idealized facial imagery with negative body-image outcomes, from ``Snapchat Dysmorphia'' \citep{ramphul2018snapchat} (patients seeking cosmetic surgery to match photo filters) to increased body dissatisfaction after exposure to more attractive social-media profiles \citep{haferkamp2011social,brown2016attractive,yang2020selfie,holland2016systematic}. If T2I models default to or exclusively generate more attractive faces, this risks reinforcing narrow beauty ideals at scale.

\paragraph{Bias in Machine Learning}
Bias has been extensively studied in large language models \citep{guo2024bias}, vision-language models \citep{chuang2023debiasing,alabdulmohsin2024clip,zhang2025joint,zhu2023debiased}, image classifiers \citep{espinosa2024efficient}, and diffusion models \citep{friedrich2023fair,kim2024training,han2026lightfair}. Most research focuses on demographic attributes such as gender, age, and ethnicity, with comparatively less attention to attractiveness \citep{gulati2024lookism}, however, recent work has begun to examine how facial attractiveness can influence machine learning systems. \citet{taylor2026algorithmic} find the LAION-Aesthetic Predictor \citep{schuhmann2022laionaesthetics} scores images of women systematically higher than images of men and LGBTQ+ individuals, and argue that reducing aesthetic quality to a scalar score is inherently prone to encoding such bias. \citet{doh2025algorithms} shows Stable Diffusion v2 associates attractiveness with trustworthiness and intelligence in portrait generation, whereas \citet{dinkar2025erasing} finds prompts with negative beauty-related traits are more likely to produce explicit images. These findings suggest attractiveness can become embedded in ML systems through both data and evaluation, but little is known about how it shapes human preferences between AI-generated images or propagates through the models trained on them.

%% file: sec/3_key_observation.tex
\section{Auditing Attractiveness Bias}
\label{sec:experiments}

\subsection{User Study}
\label{sec:key_observation}

Each human preference dataset has its own distinct approach to gathering human preference data. For example, Pick-a-Pic \citep{kirstain2023pick} presents users with two images and directs the user to ``Click the image you personally prefer..''. HPD \citep{wu2023human,ma2025hpsv3} gathers their data by asking users to rank 4 images, but instructions mix different properties, directing the user to consider ``personal aesthetic appeal'', ``text-image alignment'', and ``image quality'' when ranking images. Out of all human preference datasets considered ImageRewardDB \citep{xu2023imagereward} focuses least on personal preference or aesthetics in its user instructions. Specifically, ImageRewardDB asked users to rank 4 images according to ``...image-text alignment, high fidelity, and harmlessness.''.

As we show in Section~\ref{sec:reward_bias}, the human preference model trained on ImageRewardDB exhibits less attractiveness bias than those trained on the other datasets. One possible explanation is that instructions emphasizing ``personal preference'' or ``aesthetic appeal'' encourage users to favour more attractive faces, whereas instructions focused on ``image fidelity'' or ``image quality'' may lead to different judgments.

\begin{figure}[t]
  \centering
    \begin{tikzpicture}[x=\linewidth]
      \draw[]
        node[midway, below=0pt, font=\sffamily\bfseries, align=center] {Questions 1: Face Ranking};
    \end{tikzpicture}
    \vspace{0.3em}

    \includegraphics[width=0.75\linewidth]{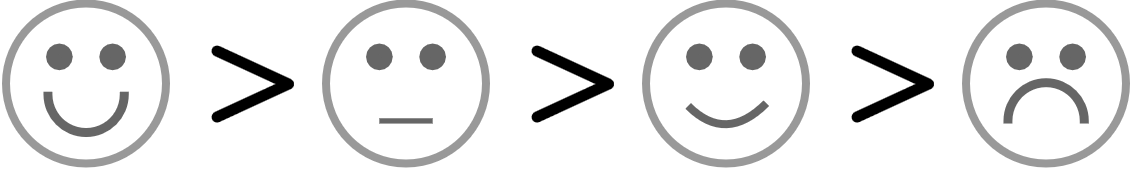}

    \vspace{0.8em}
    \begin{tikzpicture}[x=\linewidth]
      \draw[]
        node[midway, below=0pt, font=\sffamily\bfseries, align=center] {Questions 2: Face Labeling};
    \end{tikzpicture}
    \vspace{0.3em}

    \includegraphics[width=0.75\linewidth]{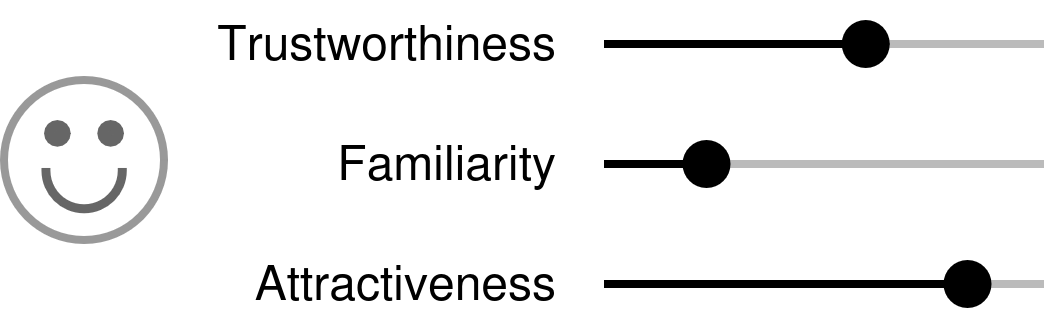}
  \caption{User study overview. Participants rank four faces by preference (Question 1) and rate three subjective attributes per face on a 0–100 scale (Question 2).}
  \label{fig:user_study}
\end{figure}

To test this, we conduct an anonymized user study on Qualtrics, recruiting 51 participants. We exclude 4 participants for revoked consent or incomplete survey responses, for a total of 47 participants. Similar to ImageRewardDB and HPD, we ask participants to rank 4 AI generated face images. Users are presented with one of two instructions:
\begin{enumerate}
    \item ``Please rank these AI generated images in terms of \textbf{image fidelity/quality}.''
    \item ``Please rank these AI generated images in terms of \textbf{your aesthetic preference}.''
\end{enumerate}
Users are asked to rank 20 sets of 4 images in this way. Users are randomly shown 1 of 5 possible sets of images (so $5\times20\times4=400$ images are utilized in the study). Question and image ordering is also randomized. Following \citep{makowski2025too}, users are then asked to rate three subjective attributes (0 - 100 scale) for each face: \textit{trustworthiness}, \textit{familiarity}, and \textit{attractiveness}. However, only half of the images are actually AI generated (using the state-of-the-art CelebA-HQ RelayDiffusion model \citep{teng2024relay}). The other half of the faces are real images taken from the CelebA-HQ dataset \citep{karras2017progressive}. Since users are led to believe all images are AI generated, they should rank images according to which they think are better generations. And since we also have a numeric score showing how attractive participants find each face, we can see if the attractiveness of a face influences its ranking. If we only consider real images, we can eliminate the possible confounder of variations in image fidelity within our ranking. \revision{We assume the real images are of comparable fidelity and inherently more realistic than generated images.} See Figure~\ref{fig:user_study} for a visual overview of the user study questions.

For each ranking of 4 images we therefore consider the two real images, referring to the higher ranked image as the \textit{favoured} face and the lower ranked images as the \textit{unfavoured} face. To determine whether participants are more likely to rank faces more highly if they find them more attractive, we formulate the following hypothesis:

\begin{tcolorbox}[colback=gray!10]

\textbf{Hypothesis 1 (Attractiveness Bias).}
\begin{align*}
H_{0}^{(1)} :& ~\text{Perceived facial attractiveness is not} \\ & ~\text{associated with image ranking.} \\
H_{A}^{(1)} :& ~\text{Images rated as more attractive are more } \\&~\text{likely to receive a higher ranking.}
\end{align*}

\end{tcolorbox}

To test Hypothesis~1 we use a one-sided Wilcoxon signed-rank test with a significance level of $\alpha=0.05$ to determine if the participant-level attractiveness difference between the \textit{favoured} and \textit{unfavoured} faces is greater than zero. To ensure independence of observations, we first average the attractiveness differences across all rankings within each participant. The Holm method is applied to control the family-wise error rate. For the ``fidelity'' and ``aesthetic'' instructions we obtain p-values of 0.0018 and 0.0006, respectively. Both are well below the significance threshold, allowing us to reject the null hypothesis and conclude that participants tend to rank the faces they find more attractive more highly under both instruction conditions. We also found that participants who took more time to complete the survey tended to show less attractiveness bias. However, the relationship was not statistically significant.

To determine whether the ``aesthetic'' wording produces more attractiveness bias than the ``fidelity'' wording we formulate the following hypothesis:

\begin{tcolorbox}[colback=gray!10]
\textbf{Hypothesis 2 (Instruction Effect).}
\begin{align*}
H_{0}^{(2)} :&~\text{The association between attractiveness and} \\&~\text{image ranking is the same under both} \\&~\text{instruction conditions.} \\
H_{A}^{(2)} :&~\text{The association between attractiveness and} \\&~\text{image ranking is stronger when participants}\\
&~\text{are instructed to rank images according to } \\&~\text{their \emph{aesthetic preference} than when instructed}\\
&~\text{to rank according to \emph{image fidelity/quality}.}
\end{align*}
\end{tcolorbox}

To test Hypothesis~2 we use a one-sided Mann--Whitney U test with a significance level of $\alpha=0.05$ to compare the participant-level attractiveness differences between the two instruction conditions. As before, we average the attractiveness differences within each participant to ensure independence of observations, \revision{but we do not apply the Holm method as this test contains a single planned comparison}. This analysis yields a p-value of 0.028, allowing us to reject the null hypothesis and conclude that the attractiveness difference is larger under the ``aesthetic'' instruction than under the ``fidelity'' instruction.

Performing the same analysis on the other two subjective attributes, \textit{trustworthiness} and \textit{familiarity}, revealed that images rated as more trustworthy also tended to receive higher rankings, although the larger p-values indicate weaker evidence for this association. In contrast, faces rated as more familiar were not significantly more likely to receive higher rankings. Furthermore, no significant difference between the two instruction conditions was found for either trustworthiness or familiarity.

To summarize, our analysis suggests that humans may have a bias toward labeling a more attractive face as better quality. Our findings also suggest that this effect can be reduced, although not entirely eliminated, by asking users to rank images in terms of ``{image fidelity/quality}'', rather than by ``{aesthetic preference}''. We next trace how this finding propagates through leaderboards (Section~\ref{sec:leaderboard}), reward models (Section~\ref{sec:reward_bias}), and generation (Section~\ref{sec:generation_bias}).

%% file: sec/4_leaderboard.tex
\subsection{Leaderboards Skew Toward Attractive Faces}
\label{sec:leaderboard}

Section~\ref{sec:key_observation} established that, among real images, individual human raters select more attractive faces as better generations. We now ask whether this bias, aggregated over many pairwise comparisons, is large enough to distort the leaderboards that these comparisons are used to build.

A popular example of an image generation leaderboard is LMArena. This platform allows a user to enter a prompt, generates two images from two separate text-to-image models, and then asks the user to pick the image they prefer. This pairwise preference data is subsequently used to calculate an Elo rating for each model. These Elo ratings determine the overall ranking of all models, where a higher score indicates a better rank.

To test whether attractiveness bias systematically skews leaderboard results, we run 10,000 Elo ranking simulations \revision{(initial Elo 1500, K-factor 64)} using the user preference data gathered from our user study. We define two competing teams: (1) less attractive, real CelebA-HQ images, and (2) more attractive, AI-generated CelebA-HQ images. For each set of four images in our study, we select the image from each category based on the participants' attractiveness labels. 
The average attractiveness labels are 34.1 and 67.9 for the real and AI-generated images, respectively. We run these simulations by randomizing the sequence of pairwise comparisons, updating the Elo ratings after each match based on the participants' winning preference. \revision{We also run the reverse selection, pairing more attractive real images against less attractive AI images (mean attractiveness 63.5 vs.\ 53.7).}

\revision{As shown in Table~\ref{tab:elo}, the more attractive AI team leads by $\approx$124 Elo points, whereas the reverse selection gives real images only a narrow lead. Sweeping the attractiveness gap (see Supplementary Material) shows Elo trending upward with it. Relative attractiveness, rather than image source, therefore largely decides the winner, suggesting preference-based leaderboards could be vulnerable to models that generate more attractive, rather than better, faces.}

\begin{table}[t]
  \scriptsize
  \centering
  \setlength{\tabcolsep}{6pt}
  \renewcommand{\arraystretch}{1.25}
  \begin{tabular}{@{}lcccc@{}}
    \toprule
    & \multicolumn{2}{c}{\textbf{AI More Attractive}} & \multicolumn{2}{c}{\textbf{Real More Attractive}} \\
    & \multicolumn{2}{c}{($\Delta = -33.8$)} & \multicolumn{2}{c}{($\Delta = +9.8$)} \\
    \cmidrule(lr){2-3} \cmidrule(l){4-5}
    \textbf{Instruction} & \textbf{Real Images} & \textbf{AI Images} & \textbf{Real Images} & \textbf{AI Images} \\
    \midrule
    Fidelity  & 1441.3 $\pm 0.6$& \textbf{1558.7} $\pm 0.6$ & 1498.6 $\pm 0.6$ & 1501.4 $\pm 0.6$\\
    Aesthetic & 1434.4 $\pm 0.6$& \textbf{1565.6} $\pm 0.6$ & \textbf{1512.1} $\pm 0.6$ & 1487.9 $\pm 0.6$\\
    All       & 1438.1 $\pm 0.6$& \textbf{1561.9} $\pm 0.6$& \textbf{1505.8} $\pm 0.6$ & 1494.2 $\pm 0.6$\\
    \bottomrule
  \end{tabular}
  \caption{\revision{Mean final Elo ratings (± standard error) of real and AI-generated CelebA-HQ images over 10000 simulated matchmaking runs using user study preferences. Left: less attractive real images against more attractive AI images. Right: the reverse selection. $\Delta$ is the difference in mean attractiveness rating (Real $-$ AI).}}
  \label{tab:elo}
\end{table}

% \begin{table}[t]
%   \footnotesize
%   \centering
%   \setlength{\tabcolsep}{8pt}
%   \renewcommand{\arraystretch}{1.25}
%   \begin{tabular}{@{}lcc@{}}
%     \toprule
%     \textbf{Instruction Type} & \textbf{Real Images} & \textbf{AI Images} \\
%     \midrule
%     Fidelity Instruction   & $1441.3 \pm 0.6$ & $\mathbf{1558.7} \pm 0.6$ \\
%     Aesthetic Instruction & $1434.4 \pm 0.6$ & $\mathbf{1565.6} \pm 0.6$ \\
%     All Responses   & $1438.1 \pm 0.6$ & $\mathbf{1561.9} \pm 0.6$ \\
%     \bottomrule
%   \end{tabular}
%   \caption{Mean Elo ratings \revision{(± standard error)} when comparing more attractive AI-generated CelebA-HQ images against less attractive, but real CelebA-HQ images. User study preference data is used to simulate randomized matchmaking between the two groups over 10000 runs. We report the simulated outcomes across ``fidelity'' instructions, ``aesthetic'' instructions, and combined responses.}
%   \label{tab:elo}
% \end{table}

%% file: sec/5_reward_bias.tex
\begin{table*}[t]
  \scriptsize
  \centering
  \setlength{\tabcolsep}{1.9pt}
  \renewcommand{\arraystretch}{1.1}
  \begin{tabular}{@{}lccccccccccc@{}}
    \toprule
    & & \multicolumn{10}{c}{\textbf{Absolute Correlation Coeff. Ranking}}\\
    \cmidrule(l){3-12}
    & & \multicolumn{2}{c}{\textbf{1st}} & \multicolumn{2}{c}{\textbf{2nd}} & \multicolumn{2}{c}{\textbf{3rd}} & \multicolumn{2}{c}{\textbf{4th}} & \multicolumn{2}{c}{\textbf{5th}}\\
    \cmidrule(lr){3-4} \cmidrule(lr){5-6} \cmidrule(l){7-8} \cmidrule(l){9-10} \cmidrule(l){11-12}
    \textbf{Reward Model} & \textbf{\makecell{Human Pref.\\Model}} & Attribute & $r$ & Attribute & $r$ & Attribute & $r$ & Attribute & $r$ & Attribute & $r$\\
    \midrule
    Aesthetic \citep{schuhmann2022laionaesthetics}   & \xmark & Blond Hair & $0.37$ & Pointy Nose & $0.35$ & Wearing Earrings & $0.34$ & Arched Eyebrows & $0.32$ & Mustache & $-0.31$ \\
    CLIP  \citep{radford2021learning} & \xmark &  Heavy Makeup & $-0.33$ & Bushy Eyebrows & $0.33$ & Wearing Lipstick & $-0.31$ & Male & $0.28$ & Wearing Necklace & $-0.27$\\
    ImageReward \citep{xu2023imagereward} & \cmark & Narrow Eyes & $-0.39$ & Wearing Necklace & $-0.34$ & Bushy Eyebrows & $0.32$ & \attr{Attractive} & $0.30$ & Mouth Slightly Open  & $-0.27$ \\
    PickScore \citep{kirstain2023pick}  & \cmark & \attr{Attractive} & $0.64$ & Young & $0.56$ & Oval Face & $0.55$ & Arched Eyebrows & $0.52$ & Wearing Lipstick & $0.51$\\
    MPS \citep{zhang2024learning}        & \cmark & \attr{Attractive} & $0.49$ & Pointy Nose & $0.49$ & Chubby & $-0.46$ & Double Chin & $-0.35$ & Arched Eyebrows & $0.32$ \\
    HPSv2.1 \citep{wu2023human}     & \cmark & \attr{Attractive} & $0.68$ & Chubby & $-0.58$ & Oval Face & $0.49$ & Bushy Eyebrows & $0.48$ & Rosy Cheeks & $0.41$\\
    HPSv3 \citep{ma2025hpsv3}      & \cmark & \attr{Attractive} & $0.52$ & Oval Face & $0.47$ & Chubby & $-0.45$ & Bushy Eyebrows & $0.38$ & Young & $0.37$\\
    \bottomrule
  \end{tabular}
  \caption{Attributes that correlate most strongly with reward model scores. We edit individual attributes in CelebA-HQ images using DiffAE. For each image the reward model scores are calculated. Correlation coefficients between reward model scores and attribute edit strengths are shown here, with the five strongest correlations being shown for each reward model.}
  \label{tab:example}
\end{table*}

\subsection{Reward Models Inherit the Bias}
\label{sec:reward_bias}

If reward models are trained on the kind of comparison data described in Section~\ref{sec:key_observation}, we would expect them to inherit the same attractiveness bias observed in individual raters. In this section, we test this directly, examining whether the human preference models used to evaluate, guide, and finetune text-to-image models are biased towards attractive faces. We also include two reward models that are \emph{not} trained on human preference data, the LAION-Aesthetics predictor \citep{schuhmann2022laionaesthetics} and CLIP \citep{radford2021learning}, as a control.

\begin{figure}[t]
  \centering
  % --- image row: all five identical size; frame drawn INSIDE base image ---
  \includegraphics[width=0.188\columnwidth]{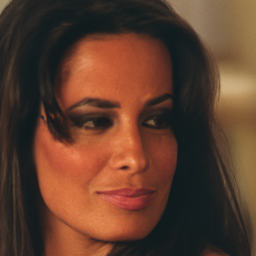}\hfill
  \includegraphics[width=0.188\columnwidth]{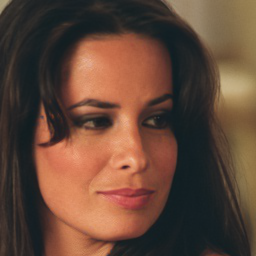}\hfill
  \begin{tikzpicture}[inner sep=0, outer sep=0]
    \node[inner sep=0] (img) {\includegraphics[width=0.188\columnwidth]{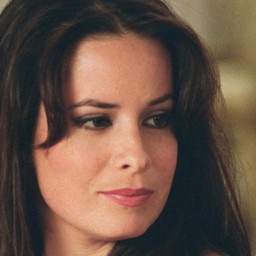}};
    \draw[baseblue, line width=1pt]
      ([xshift=0.5pt,yshift=0.5pt]img.south west) rectangle
      ([xshift=-0.5pt,yshift=-0.5pt]img.north east);
  \end{tikzpicture}\hfill
  \includegraphics[width=0.188\columnwidth]{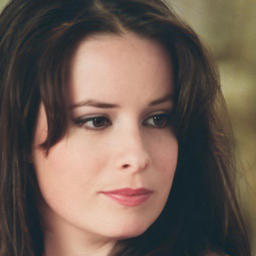}\hfill
  \includegraphics[width=0.188\columnwidth]{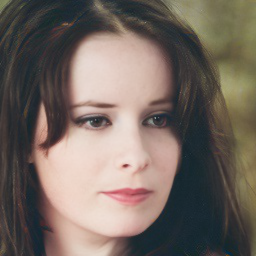}\\[3pt]
  % --- arrows ---
  \begin{tikzpicture}[x=\columnwidth]
    \draw[-{Stealth[length=2mm]}, line width=0.8pt, arrowgray]
      (0.394,0) -- (0.002,0)
      node[midway, below=1pt, font=\sffamily\footnotesize\bfseries] {$-$ Pale Skin};
    \node[font=\sffamily\footnotesize\bfseries, baseblue] at (0.5,-0.15) {Base};
    \draw[-{Stealth[length=2mm]}, line width=0.8pt, arrowgray]
      (0.606,0) -- (0.998,0)
      node[midway, below=1pt, font=\sffamily\footnotesize\bfseries] {$+$ Pale Skin};
  \end{tikzpicture}
  \caption{Example Diffusion Autoencoder editing of a CelebA-HQ image. Diffusion Autoencoder used to edit the ``Pale Skin'' attribute, with the original unedited image showing in the middle.}
  \label{fig:diff_ae_example}
\end{figure}

\paragraph{Evaluation on Synthetic Images}

\begin{figure*}[t]
  \begin{minipage}[c]{0.74\textwidth}
      \centering
      \includegraphics[width=1.0\columnwidth]{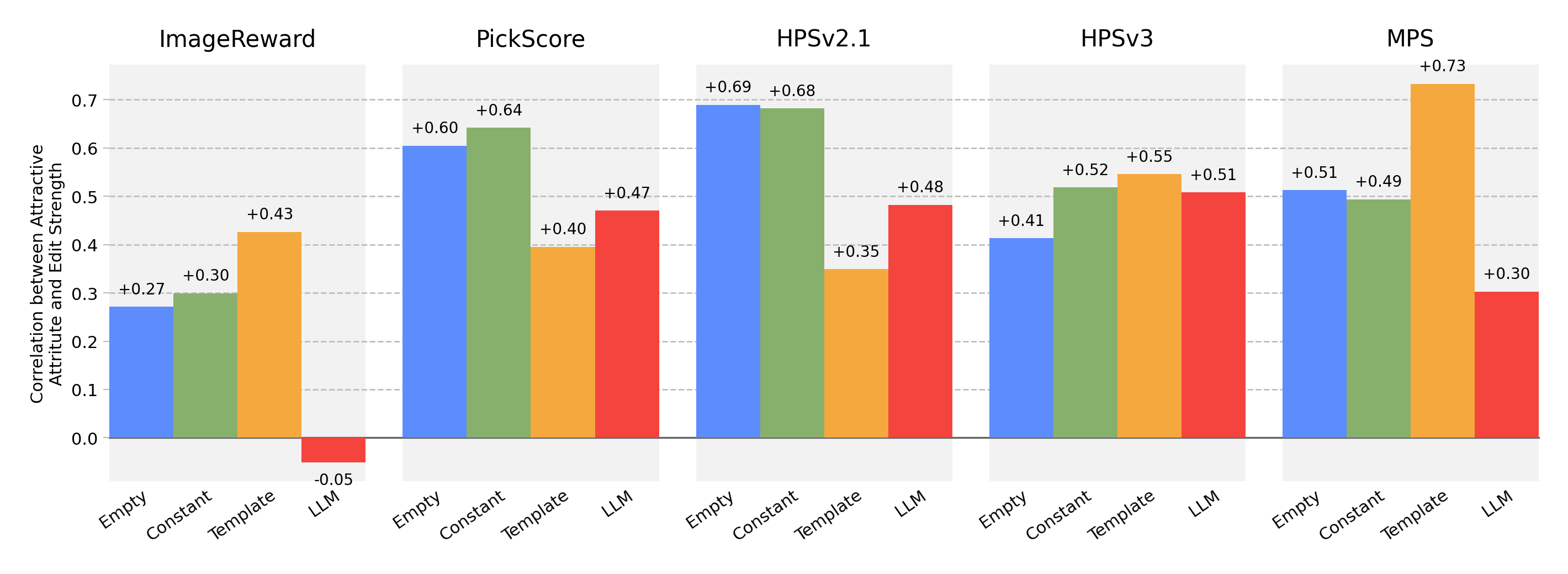}
      \caption{Caption strategies on synthetic CelebA-HQ images with the \textit{attractive} attribute edited using Diffusion Autoencoder. Bars show the correlation coefficients between the edit strength and the reward model score. For each reward model (ImageReward, PickScore, HPSv2.1, HPSv3, MPS) we consider four captioning strategies: empty, constant, template-based, and LLM-generated.}
      \label{fig:cap_strat}
  \end{minipage}
  \hfill
  \begin{minipage}[c]{0.24\textwidth}
  \footnotesize
  \centering
  \setlength{\tabcolsep}{0.5pt}
  \renewcommand{\arraystretch}{1.1}
  \begin{tabular}{@{}lcc@{}}
    \toprule
    \textbf{Ranking} & \textbf{Attribute} & \textbf{$\lambda_{L}$} \\
    \midrule
    1st & \attr{Attractive}    & 0.63 \\
    2nd & Wearing Hat & 0.25 \\
    3rd & Oval Face & 0.25 \\
    4th & 5 O'Clock Shadow & 0.22 \\
    5th & Bushy Eyebrows & 0.22 \\
    \bottomrule
  \end{tabular}
  \captionof{table}{Top five most predictive CelebA-HQ facial attributes for HPSv3 scores. Attributes are ranked by the LASSO regularization strength $\lambda_{L}$ at which they are finally removed from the model.}
  \label{tab:lasso}
  \end{minipage}
\end{figure*}

We start by experimenting on CelebA-HQ because this dataset has a binary label for \textit{attractive} \citep{liu2015faceattributes}. Specifically, we take 100 base CelebA-HQ images of resolution 256 by 256 and use a pretrained Diffusion Autoencoder \citep{preechakul2022diffusion} to change individual attributes in isolation. We use 4 randomized edit magnitudes between $-0.5$ and $0.5$. For each base image this produces 4 images for each of CelebA-HQ's 40 attributes, for a total of $4\times40\times100=16000$ synthetic images. For an example of the image edits generated, please see Figure~\ref{fig:diff_ae_example}.

We then run the following reward models on, and get scores for, each image: LAION-Aesthetics predictor \citep{schuhmann2022laionaesthetics}, CLIP \citep{radford2021learning}, ImageReward \citep{xu2023imagereward}, PickScore, \citep{kirstain2023pick}, MPS, \citep{zhang2024learning}, HPSv2.1 \citep{wu2023human}, and HPSv3 \citep{ma2025hpsv3}. For models that require a caption we use the generic text ``a photo of a face''. We are then interested in the following question: \textit{Does increasing the prominence of certain facial attributes result in higher rewards?}

To answer this, we calculate the correlations between the rewards given and the edit strengths. Table~\ref{tab:example} shows which of the 40 attributes in the CelebA-HQ dataset most strongly correlate with each reward. Critically, our two non-preference control models behave differently from the five preference models: there is no discernible pattern for which attributes most strongly correlate with the LAION-Aesthetics predictor's scores, and CLIP's strongest correlations run through attributes associated with women (e.g. \textit{wearing earrings}, \textit{heavy makeup}) rather than attractiveness. In contrast, \textit{attractive} is the attribute that most strongly correlates with scores from PickScore, MPS, HPSv2.1, and HPSv3, all of which are human preference models. Importantly, many of these models are finetuned from CLIP, suggesting that this pattern only emerges from human preference alignment.

To examine the impact of the prompt on these correlations, we repeat this analysis using four different captioning strategies.

\begin{enumerate}
\item An \textbf{empty} caption ``'' is used for all images.
\item A \textbf{constant} caption ``a photo of a face'' is used for all images.
\item A \textbf{template} caption is used based on the edited attribute; for example, for the \textit{male} attribute, we use the captions ``a photo of a man's face'' or ``a photo of a woman's face''.
\item An \textbf{LLM} generated caption is used for each image using the Gemma 4 \citep{team2026gemma} model.
\end{enumerate}

Figure~\ref{fig:cap_strat} shows how these different captioning strategies impact the correlations between edit strength and the \textit{attractive} attribute. Compared to our prior results using a constant caption, moving to LLM generated captions tends to reduce, but not eliminate the correlation. Importantly, attractiveness bias persists regardless of caption used. The one notable exception to this is ImageReward, which already has the lowest correlation, and for which an LLM based caption eliminates the bias. As discussed in Section~\ref{sec:key_observation}, ImageReward is also the reward model whose annotation instructions place the least emphasis on personal or aesthetic preference, which offers a plausible explanation for why it is comparatively less affected by attractiveness bias.

Although these results suggest that human preference models may have inherited a bias for more attractive faces, it is plausible that some attribute may produce better quality edits due to imbalances in the training data. Next, we provide additional analysis using real image datasets, where image fidelity should remain constant, thereby eliminating any DiffAE artefacts as a possible confounder.

\paragraph{Evaluation on Real Images}

\begin{figure}[t]
\centering
\includegraphics[width=1.0\columnwidth]{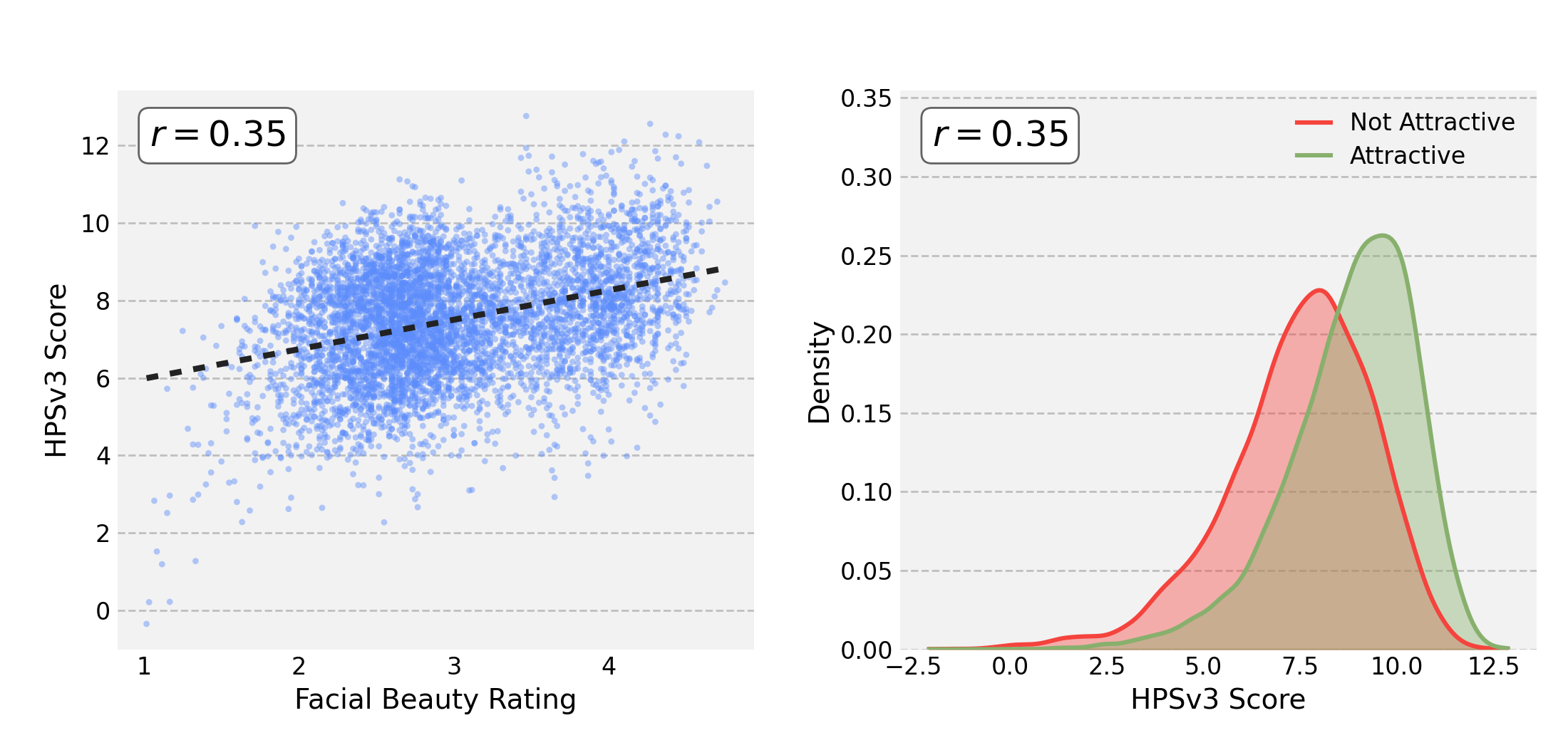}
\caption{Left: the SCUT-FBP5500 dataset's facial beauty labels against HPSv3
scores. Right: distribution of HPSv3 scores on real CelebA-HQ images, separated
by the binary \textit{attractive} label.}
\label{fig:beauty_hps3}
\end{figure}

To analyse human preference models on real images we must consider the domains they were trained on. The majority of human preference models we consider are trained on AI generated images exclusively, making real images out-of-distribution. The one exception is HPSv3, which was trained on a mixture of both AI generated images and real images. We therefore focus primarily on this model when considering real images.

We evaluate on two datasets: SCUT-FBP5500 \citep{liang2017SCUT} and CelebA-HQ \citep{liu2015faceattributes,karras2017progressive}. Both datasets contain images of faces and accompanying labels for attractiveness. Specifically, CelebA-HQ uses binary labels and SCUT-FBP5500 uses continuous labels from 1 to 5. Evaluation uses a constant caption. Figure~\ref{fig:beauty_hps3} (left) shows the HPSv3 score on real SCUT-FBP5500 images plotted against their facial beauty labels, finding a correlation coefficient of $r=0.35$. Figure~\ref{fig:beauty_hps3} (right) shows the distribution of HPSv3 scores on real CelebA-HQ images for attractive and unattractive faces. We also get a correlation coefficient of $r=0.35$ for this dataset. A positive correlation still existing for real images suggests that the bias toward attractive faces found previously cannot be attributed to image quality variation in synthetic images.

We also consider which facial attributes are most predictive of human preference scores. To do so, we perform a LASSO regression on CelebA-HQ attributes \revision{(standardized before regression)}, using the HPSv3 score as the target variable. By steadily increasing the regularization strength $\lambda_{L}$, we can identify the most influential features by monitoring which attributes are retained the longest before being eliminated from the model.

The results in Table~\ref{tab:lasso} indicate that the \textit{attractive} attribute is the strongest predictor of the HPSv3 score, persisting up to a regularization strength of $\lambda_{L} = 0.63$. \revision{This suggests that attractiveness carries substantial predictive information about HPSv3 scores.}

%% file: sec/6_generation_bias.tex
\begin{figure*}[t]
  \centering
  \begin{minipage}[c]{0.60\textwidth}
    \centering

    \setlength{\tabcolsep}{0pt}
    \renewcommand{\arraystretch}{0}
    \newlength{\rowlabelwidth}
    \setlength{\rowlabelwidth}{14pt}
    \newlength{\imgw}
    \setlength{\imgw}{\linewidth}
    \addtolength{\imgw}{-\rowlabelwidth}
    \setlength{\imgw}{0.16666\imgw}

    \begin{tabular}{@{}m{\rowlabelwidth}@{}*{6}{>{\centering\arraybackslash}m{\imgw}}@{}}
      \centering\rotatebox{90}{\sffamily\fontsize{8}{9.6}\selectfont\bfseries\textcolor{arrowgray}{PixArt-$\alpha$}} &
      \includegraphics[width=\imgw]{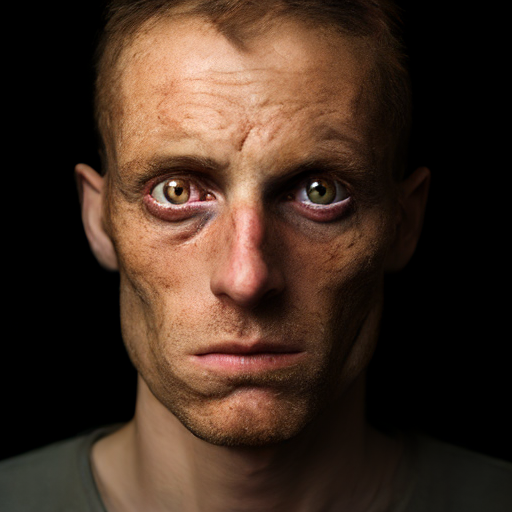} &
      \includegraphics[width=\imgw]{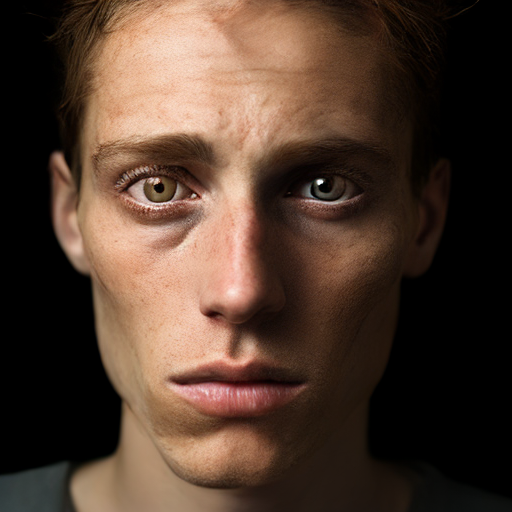} &
      \includegraphics[width=\imgw]{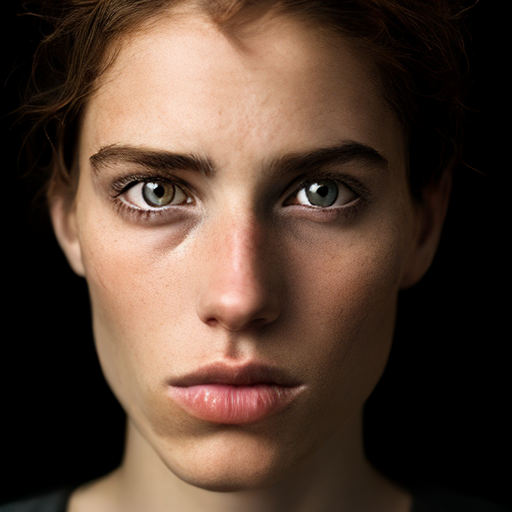} &
      \includegraphics[width=\imgw]{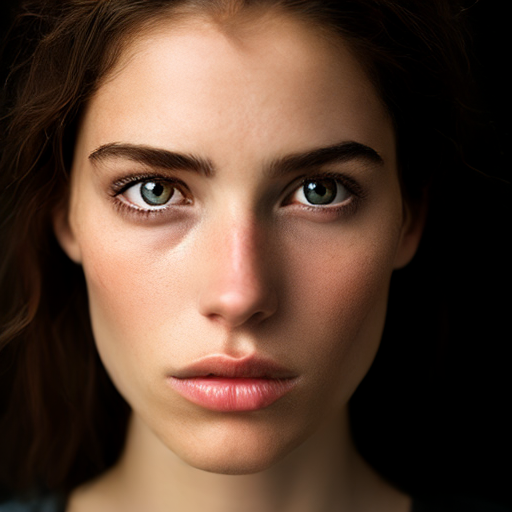} &
      \includegraphics[width=\imgw]{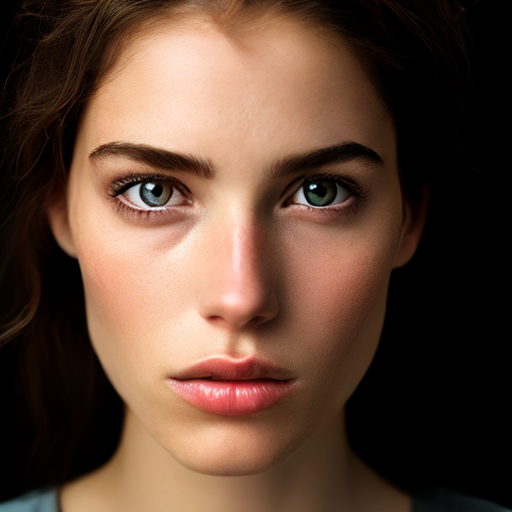} &
      \includegraphics[width=\imgw]{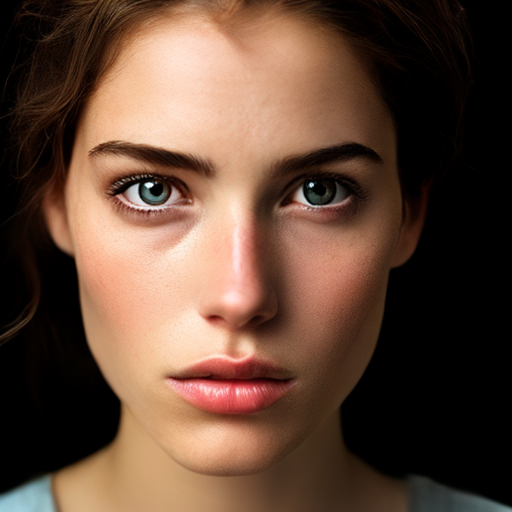} \\
      \centering\rotatebox{90}{\sffamily\fontsize{8}{9.6}\selectfont\bfseries\textcolor{arrowgray}{SD-Turbo}} &
      \includegraphics[width=\imgw]{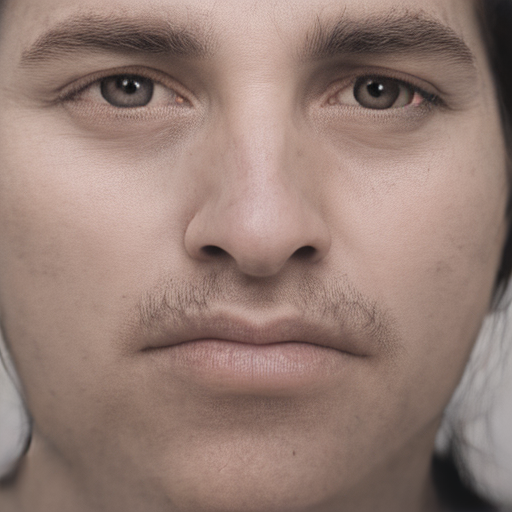} &
      \includegraphics[width=\imgw]{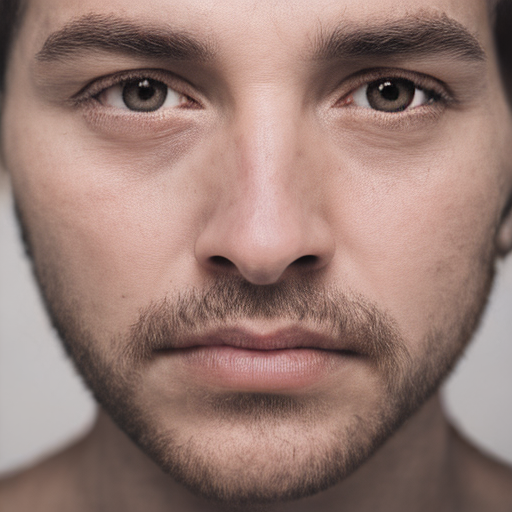} &
      \includegraphics[width=\imgw]{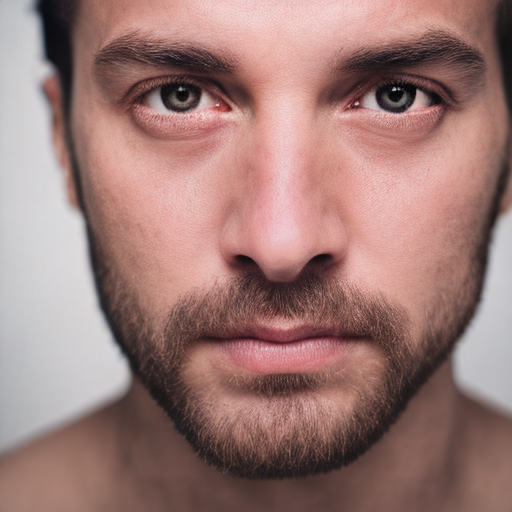} &
      \includegraphics[width=\imgw]{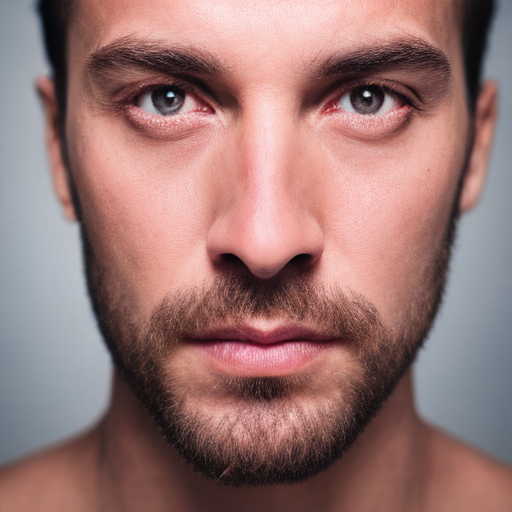} &
      \includegraphics[width=\imgw]{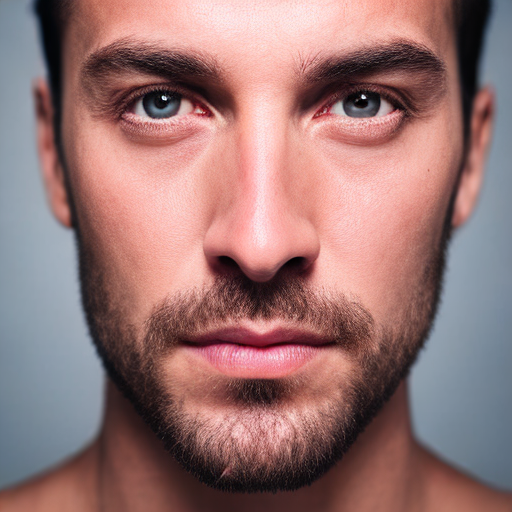} &
      \includegraphics[width=\imgw]{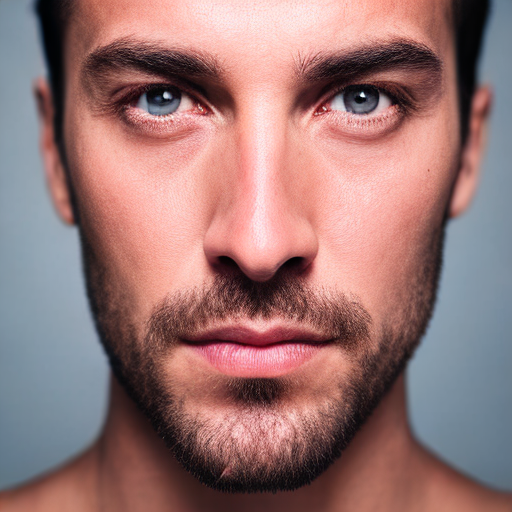} \\
    \end{tabular}

    \vspace{0.5em}

    \newlength{\imgtotalw}
    \setlength{\imgtotalw}{6\imgw}
    \noindent\hspace*{\rowlabelwidth}%
    \begin{tikzpicture}[x=\imgtotalw]
      \draw[-{Stealth[length=2mm]}, line width=0.8pt, arrowgray]
        (0.0,0.0) -- (1.0,0.0)
        node[midway, below=2pt, font=\sffamily\footnotesize\bfseries, align=center]
        {$+$ Human Preference \\ ~Alignment Steps};
    \end{tikzpicture}

    \caption{Human preference guided generation for the prompt ``a photo of an unattractive face'' using ReNO with HPSv2.1 as the reward model. Left to right, images are aligned for $[0, 4, 8, 12, 16, 20]$ steps. Each row shows the impact of optimizing the human preference score on PixArt-$\alpha$ (top) and SD-Turbo (bottom).}
    \label{fig:reno_results}
  \end{minipage}%
  \hfill
  \begin{minipage}[c]{0.38\textwidth}
    \centering
    \small
    \setlength{\tabcolsep}{4pt}
    \renewcommand{\arraystretch}{1.1}
    \begin{tabular}{@{}lcccc@{}}
      \toprule
      \textbf{Model} & \textbf{ReNO} & \textbf{Attract.} & \textbf{HPSv2.1} \\
      \midrule
      \multirow{2}{*}{PixArt-$\alpha$} & \xmark & \textbf{0.15}$\pm$0.14 & 0.284$\pm$0.009 \\
                                        & \cmark & 0.87$\pm$0.13 & \textbf{0.348}$\pm$0.014 \\
      \midrule
      \multirow{2}{*}{SD-Turbo} & \xmark & \textbf{0.19}$\pm$0.16 & 0.241$\pm$0.018 \\
                                 & \cmark & 0.66$\pm$0.26 & \textbf{0.330}$\pm$0.022 \\
    \midrule
    \multirow{2}{*}{SDXL-Turbo} & \xmark & \textbf{0.49}$\pm$0.19 & 0.282$\pm$0.013 \\
                                 & \cmark & 0.69$\pm$0.20 & \textbf{0.309}$\pm$0.013 \\
      \bottomrule
    \end{tabular}
    \captionof{table}{Average predicted attractiveness and HPSv2.1 scores \revision{(± standard deviation)} for images before and after ReNO human preference optimization using the prompt ``a photo of an unattractive face''. 20 steps are used for alignment with HPSv2.1 as the reward model.}
    \label{tab:reno_model_comparison}
  \end{minipage}
\end{figure*}

\begin{figure}[t]
  \centering
  % --- SUBFIGURE A: Preference-guided Generation Diagram ---
  \includegraphics[width=0.75\linewidth]{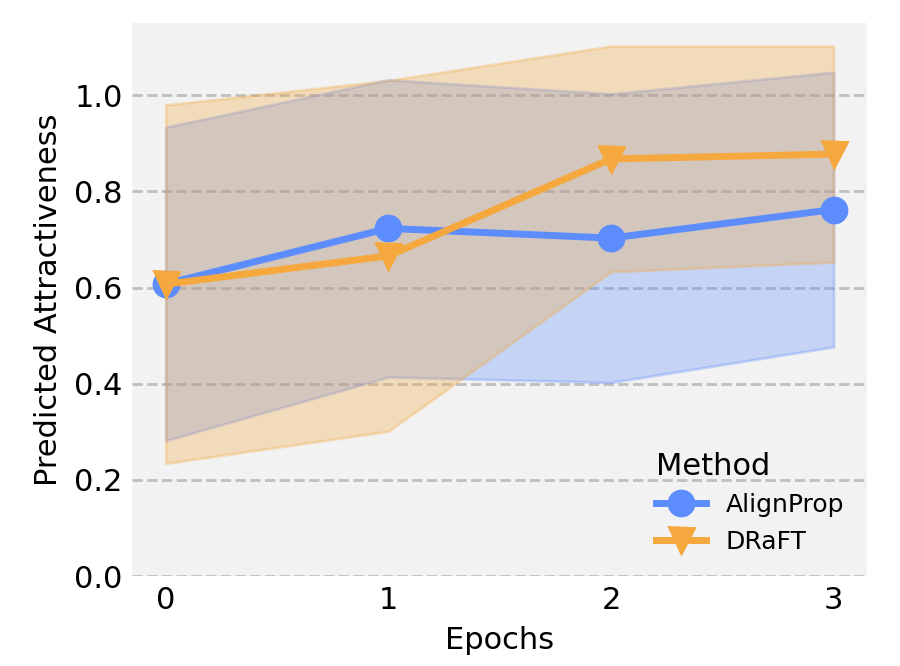}
  \caption{Human preference finetuning of Stable Diffusion v1.5 and a DDPM model using AlignProp and DRaFT, respectively. Each series shows the mean predicted facial attractiveness at each epoch of training \revision{(± standard deviation)}. Epoch 0 is the base mode (i.e. no finetuning).}
  \label{fig:finetuning_hps}
\end{figure}

\subsection{Preference Aligned Generation}
\label{sec:generation_bias}

Section~\ref{sec:reward_bias} establishes that facial attractiveness is both highly correlated with and highly predictive of human preference scores, but this does not yet establish a causal relationship. In this section, we therefore apply human preference models to guide the generation process itself and measure the impact of this on facial attractiveness. To measure facial attractiveness, we finetune DINOv3 \citep{simeoni2025dinov3} on CelebA-HQ labels from 0 (least) to 1 (most attractive), and use the resulting model to predict attractiveness in the following experiments (see Supplementary Material for details).

To investigate how finetuning diffusion models based on human preferences affects the attractiveness of generated faces, we consider two approaches. First, we finetune Stable Diffusion v1.5 using AlignProp \citep{prabhudesai2023aligning}, with HPSv2.1 as the reward model and prompts sampled from HPD. Second, we finetune a CelebA-HQ DDPM model \citep{ho2020denoising} using DRaFT-K \citep{clark2024directly} ($K=1$), with PickScore as the reward model and a constant prompt strategy. Both models are trained for three epochs, and after each epoch we generate 5000 images using 50 inference steps. For Stable Diffusion v1.5, we use prompts from the CelebA-Dialog dataset \citep{jiang2021talkedit} and a guidance scale of 5.0. Figure~\ref{fig:finetuning_hps} shows the average facial attractiveness of the images generated at each stage of finetuning. For both models, attractiveness \revision{trends upwards} %steadily increases 
as training progresses, suggesting that closer alignment with the human preference model leads to more attractive generated faces. \revision{The standard deviation of predicted attractiveness also decreases, although this partly reflects the predicted attractiveness saturating. The demographic distributions of generated faces provide a more direct measure of diversity (see Supplementary Material); after finetuning, generated faces become noticeably younger and more female.}
%At the same time, the standard deviation of attractiveness decreases, indicating a reduction in the diversity of facial attractiveness. The Supplementary Material shows further qualitative results where we observe that while this process does seem to improve image quality, it also seems to push all generated faces towards a westernized ideal of beauty (i.e. young, female, and light skinned).

Finally, we consider the impact of human preference guided generation using ReNO \citep{eyring2024reno} with the prompt ``a photo of an unattractive face''. ReNO optimizes the initial noise in a text-to-image model using a given reward model. We use 20 optimization steps, HPSv2.1 as the reward model, and consider PixArt-$\alpha$ \citep{chen2024pixart}, SD-Turbo \citep{sauer2024adversarial}, and SDXL-Turbo as our T2I models. ReNO is used to optimize 5000 images for each model.  As shown in Table~\ref{tab:reno_model_comparison}, PixArt-$\alpha$ and SD-Turbo are correctly able to render the target ``unattractive face'', however, aligning our outputs with HPSv2.1, which should push images toward better image-text alignment, leads instead towards more attractive faces. This is also shown visually in Figure~\ref{fig:reno_results}. This suggests that human preference alignment can increase facial attractiveness instead of improving text-image correspondence. If human preference alignment removes the ability to generate unattractive faces, this poses a problem for user control similar to that observed in \citep{guo2025position}. See Supplementary Material for further analysis on different prompts and the interaction with race, gender, and age. 

%% file: sec/7_mitigations.tex
\section{Mitigation}
\label{sec:mitigation}

Having established that facial attractiveness can strongly distort human preference, we now \revision{explore} bias mitigation. In particular, we focus on human preference models, whose continuous scalar scores offer a valuable opportunity to apply mathematically tractable, information-theoretic debiasing techniques. \revision{We propose the following target behaviour: preference scores should depend on attractiveness only when the prompt explicitly requests it. Our evaluation prompts do not mention attractiveness, so here the preference scores $S$ should be independent of attractiveness $A$.}

\revision{We therefore design a loss function that minimizes the mutual information between $A$ and $S$.} If the mutual information $\mathbb{I}(A;S)=0$, then knowing the attractiveness of a face provides no information about its resulting score, thereby removing the skew toward attractive faces. Kernel density estimation is used to calculate mutual information. However, we also want the human preference scores to change as little as possible so that they retain their useful signal. Our loss function is therefore:
\begin{equation}
\mathcal{L}=\|S-\hat{S}\|^2+\lambda \mathbb{I}(A;\hat{S}),
\end{equation}
where $\hat{S}$ is our adjusted (unbiased) score and $\lambda$ is a scalar that trades off the mutual information term against the score similarity term. We cannot change the attractiveness of the faces, as \revision{this is a subjective attribute of the images}. Instead, we adjust the score to remove its dependence on facial attractiveness. We therefore define a simple residual function for finding $\hat{S}$ that can be quickly adapted to new domains: 
$\hat{S}=S + \beta_0A + \beta_1$,
where $\beta$ are the learned parameters.

\begin{figure}[t]
\centering
\includegraphics[width=1.0\columnwidth]{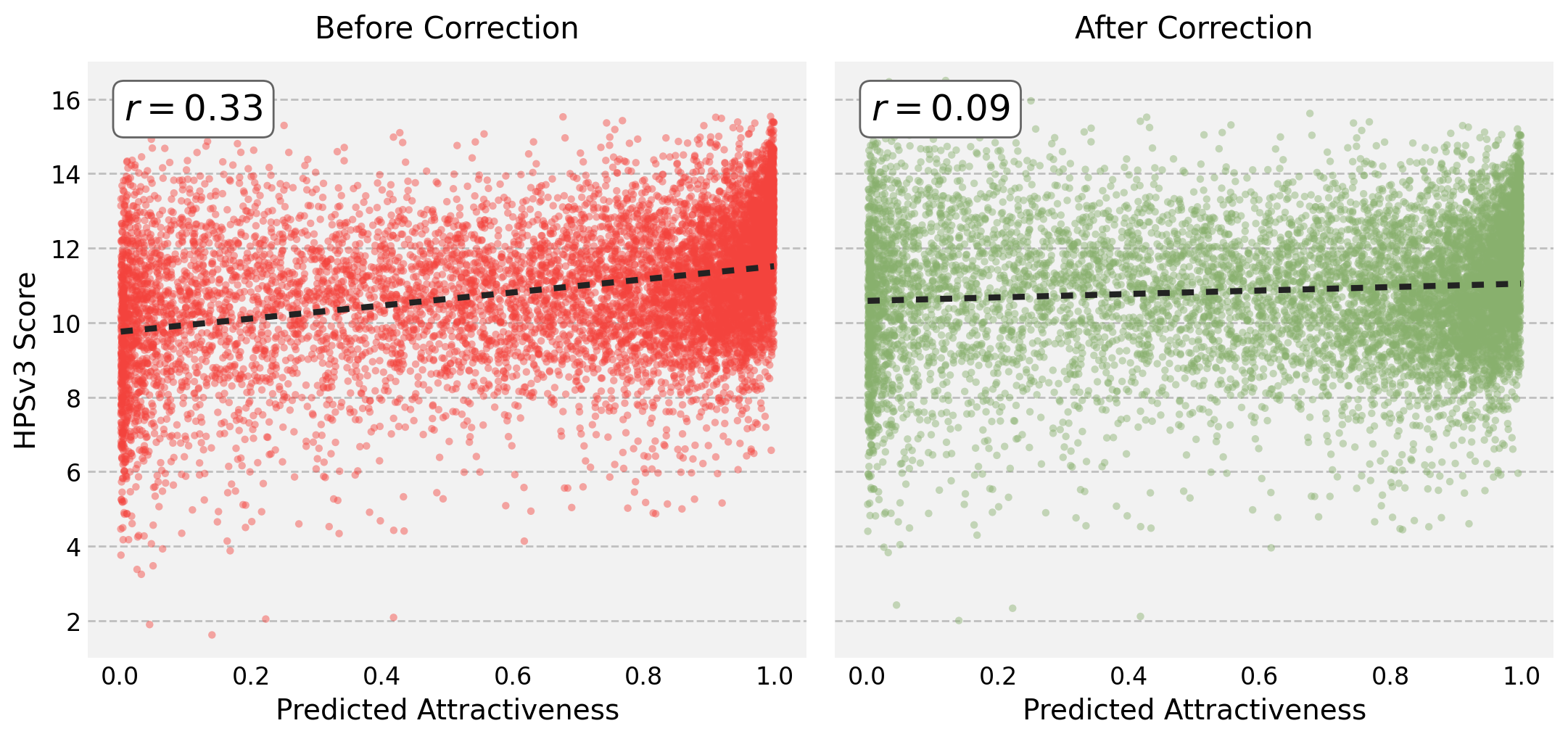}
\caption{Left: HPSv3 scores vs. facial attractiveness across 9,000 Stable Diffusion v3.5 outputs. Right: Scores after applying our debiasing approach.}
\label{fig:debiased_hpsv3}
\end{figure}

\begin{table}[t]
  \scriptsize
  \centering
  \setlength{\tabcolsep}{4pt}
  \renewcommand{\arraystretch}{1.1}
  \begin{tabular}{@{}lcccccccc@{}}
    \toprule
    & \multicolumn{2}{c}{\textbf{PickScore}} & \multicolumn{2}{c}{\textbf{HPSv2.1}} & \multicolumn{2}{c}{\textbf{HPSv3}} & \multicolumn{2}{c}{\textbf{MPS}}\\
    \cmidrule(lr){2-3} \cmidrule(lr){4-5} \cmidrule(lr){6-7} \cmidrule(l){8-9}
    \textbf{T2I Model} & Orig. & Adj. & Orig. & Adj. & Orig. & Adj. & Orig. & Adj.\\
    \midrule
    FLUX.2 \citep{flux-2-2025}  & 0.19 & \textbf{0.03} & 0.10 & \textbf{-0.01} & 0.09 & \textbf{-0.01} & 0.07 & \textbf{0.00} \\
    LongCat \citep{LongCat-Image} & 0.24 & \textbf{0.03} & 0.26 & \textbf{0.06}  & 0.25 & \textbf{0.11}  & 0.21 & \textbf{0.07} \\
    SD1.5 \citep{rombach2022high}   & 0.13 & \textbf{0.04} & 0.24 & \textbf{0.08}  & 0.35 & \textbf{0.09}  & 0.17 & \textbf{0.07} \\
    SD3.5 \citep{esser2024scaling} & 0.23 & \textbf{0.08} & 0.29 & \textbf{0.07}  & 0.33 & \textbf{0.09}  & 0.17 & \textbf{0.02} \\
    Z-Image \citep{team2025zimage}  & 0.25 & \textbf{0.11} & 0.26 & \textbf{0.10}  & 0.17 & \textbf{0.04}  & 0.31 & \textbf{0.00} \\
    \bottomrule
  \end{tabular}
  \caption{Correlation between predicted attractiveness and reward model scores before and after applying our debiasing approach. \textbf{Orig.} is the raw correlation; \textbf{Adj.} is the correlation after debiasing.}
  \label{tab:mi_correction}
\end{table}

To probe the performance of our approach, we generate a suite of 50000 images using a variety of text-to-image models and prompts from the CelebA-Dialog dataset. Specifically, we generate 10000 images with each of Stable Diffusion v1.5 \citep{rombach2022high}, Stable Diffusion v3.5 \citep{esser2024scaling}, Z-Image-Turbo \citep{team2025zimage}, LongCat-Image \citep{LongCat-Image}, and FLUX.2 Klein 9B Base \citep{flux-2-2025}. We then calculate scores using the human preference models most susceptible to attractiveness bias, allowing us to evaluate our debiasing approach under the strongest observed bias. The prompts used for the human preference models match those used for image generation.

We evaluate the effect of our debiasing approach on human preference scores. For each text-to-image model, we calibrate the $\beta$ parameters using 1000 images and evaluate on the remaining 9000 images, with $\lambda=8\operatorname{var}(S)$. As an example, Figure~\ref{fig:debiased_hpsv3} shows the effect of our approach on HPSv3 for images generated with Stable Diffusion v3.5. The correlation between predicted facial attractiveness and human preference score is significantly reduced from 0.33 to 0.09, while maintaining a similar score range and making only minimal changes to the scores. Table~\ref{tab:mi_correction} extends this analysis to other human preference and text-to-image models. Our approach consistently reduces the correlation between attractiveness and preference score, reducing moderate correlations to weak and weak correlations to near zero. This suggests that removal of the attractiveness bias from reward models is a tractable problem.

%% file: sec/8_dicussion.tex
\section{Discussion}
\label{sec:discussion}

The attractiveness halo effect refers to a cognitive bias where attractive people are assumed to possess other positive attributes, such as kindness, intelligence, or success \citep{dion1972beautiful,gulati2024beautiful}. We find that the attractiveness halo effect also occurs when gathering human preference data for AI images, resulting in users ranking more attractive faces higher than less attractive faces. Our results suggest the bias permeates the T2I ecosystem, affecting leaderboards, preference models, and preference aligned generation alike. Our Elo simulations show that attractiveness \revision{can skew rankings, } placing synthetic images above real ones and leaving leaderboards vulnerable to reward hacking by models that generate more attractive rather than better faces.

Our findings may also have broader implications for body image since exposure to attractive faces in media has been linked to body image concerns \citep{perloff2014social,brown2016attractive,yang2020selfie,isakowitsch2022augmented}. Section~\ref{sec:generation_bias} shows preference alignment both raises facial attractiveness and narrows its diversity. As AI-generated imagery proliferates on social media, preference aligned models risk reproducing a narrow beauty ideal at scale. 

Finally, Section~\ref{sec:mitigation} shows the bias can be removed from reward scores post-hoc, but further work is needed to prevent it appearing in preference data at all. Our results suggest wording matters: asking users to rank by image fidelity produces less biased responses than asking them to rank by aesthetic preference. Attractiveness bias also decreased the longer users took to complete the survey, although a larger sample is likely needed to prove the influence definitively. The comparatively low bias in ImageReward \citep{xu2023imagereward} makes its collection process of particular interest: before ranking, ImageRewardDB users write text descriptions and give numerical quality ratings, which together with its overall instructions may prime users to focus on image quality and encourage them to spend more time considering each image.  Promising directions for future work therefore include experimenting with wording, imposing time limits, and using multi-dimensional annotation.

%However, prior work suggests that the effects of attractive faces on body image may be stronger when the faces are familiar, potentially limiting the harm caused by AI-generated imagery. 

%% file: sec/9_conclusion.tex
\section{Conclusion}
\label{sec:conclusion}

In this work, we show that attractive faces can skew human preference data for AI-generated imagery. Participants tend to rate images containing more attractive faces as better, even when image fidelity is equivalent or lower. We trace the impact of this bias through the text-to-image ecosystem, showing that it can skew human preference models, distort leaderboard rankings, and bias preference aligned generation. Finally, we propose a compute-efficient approach for removing this bias from human preference scores using a mutual information minimization objective. We hope this work encourages the T2I community to refine its evaluation practices to better account for such cognitive biases.

%\paragraph{Limitation \& Future Work} Future work should focus on mitigating this bias at the data collection stage. Our results suggest that instruction tuning could be an effective way to reduce this bias, while a larger-scale user study may shed further light on the relationship between labeling time and attractiveness bias. Finally, the weaker effect of trustworthiness suggests that attractiveness is not the only facial attribute that can bias preference data. Future work should therefore investigate what other biases are present in human preference data and how they can be mitigated.

%% file: appendix.tex
\maketitle

\section{Further Discussion}
\subsection{Societal Impact} 

The bias we identify has consequences beyond leaderboards: Section 3.4 shows that preference aligned generation raises facial attractiveness while narrowing its diversity, shifting outputs toward younger, female, and lighter-skinned faces and removing users' ability to generate anything else. At scale this risks reproducing a narrow beauty ideal in systems that increasingly mediate how people see faces, with documented links to body-image harm \citep{haferkamp2011social,brown2016attractive,yang2020selfie,holland2016systematic}. Our own methodology carries risks we state plainly: we finetune a facial attractiveness predictor and use a FairFace-trained demographic classifier, both of which are open to misuse in ranking, filtering, or profiling people, and both of which encode the judgements of particular annotator populations rather than any objective property of faces. We treat them as measurement instruments for a bias audit rather than as contributions. Finally, our mitigation is only a partial remedy. It removes the dependence of reward scores on attractiveness post-hoc but inherits the predictor's own biases and does not stop the bias entering preference data at source, which is why we regard the finding that instruction wording measurably changes recorded bias as the more durable result.

\revision{\subsection{Limitations}}

\revision{Several limitations warrant caution when interpreting our findings. First, in the user study, participants rated attractiveness after ranking the same faces, so their attractiveness ratings may have been influenced by their earlier preferences. Furthermore, we assume that real photographs are generally of higher image quality than AI-generated images, however, if a real image is especially blurred (as a very small number of CelebA image are) or corrupted with noise, this may not always hold true. In our case, attribute labels suggest real images used in the user study were not blurred. Directly matching image quality across individual real and generated images is challenging, as common metrics such as FID and IS are primarily designed for evaluating distributions of images rather than the quality of individual images. Comparing real photographs with generated images therefore provides a useful, albeit imperfect, way to investigate whether attractiveness influences preferences when image sources differ. See Section~\ref{sec:appendix_user_study_details}} for further analysis around this assumption.

\revision{Similarly, our Elo experiment uses selected comparisons between real faces and generated faces to illustrate how attractiveness bias could affect leaderboard rankings. It represents constructed scenarios rather than a direct measurement of distortion in existing leaderboards. The results should therefore be interpreted as demonstrating a potential vulnerability rather than quantifying the extent of leaderboard distortion in practice.}

\revision{It is also important to note that the DiffAE editing procedure may inadvertently alter attributes correlated with attractiveness, as well as other aspects of image appearance and quality. Although the example in Figure 3 (in the main body of the paper) suggest that the intended edits can be achieved without severe visible entanglement, we cannot guarantee that attractiveness is the only property that changes. This means that the attribute-editing experiments cannot independently establish that attractiveness alone accounts for the observed preference differences. However, the real-image experiments provide complementary evidence without relying on DiffAE-generated edits, reducing our dependence on any single experimental approach.}

\revision{Taken together, our experiments provide converging evidence that facial attractiveness is associated with human preferences for text-to-image outputs and may influence reward-model scores and leaderboard rankings, but no individual experiment conclusively isolates attractiveness from all potential confounding factors. Further work using quality-matched comparisons, more exact text-to-image leaderboard setups, and broader sampling of images and participants would help establish the magnitude of these effects.}

\begin{figure*}[t]
  \centering
  \includegraphics[width=1.99\columnwidth]{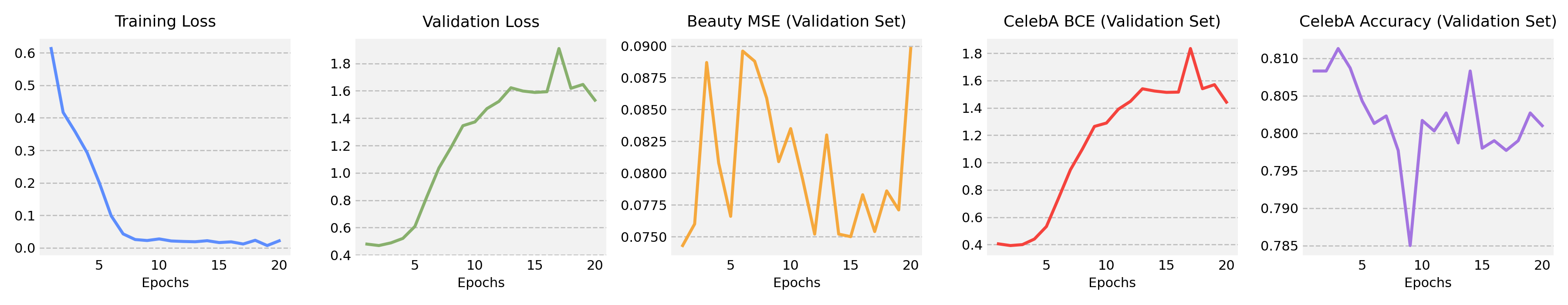}
  \caption{\revision{Loss plots and accuracy of our facial attractiveness predictor over 20 epochs. From left to right the subplots show: total training loss, total validation loss, MSE loss on the validation set of SCUT-FBP5500, BCE loss on the validation set of CelebA, and accuracy of binary attractiveness prediction on the validation set of CelebA.}}
  \label{fig:beauty_loss_plots}
\end{figure*}

\begin{figure}[t]
  \centering

  % --- SUBFIGURE A: Preference-guided Generation Diagram ---
  \begin{subfigure}{\columnwidth}
    \centering
    \includegraphics[width=0.8\linewidth]{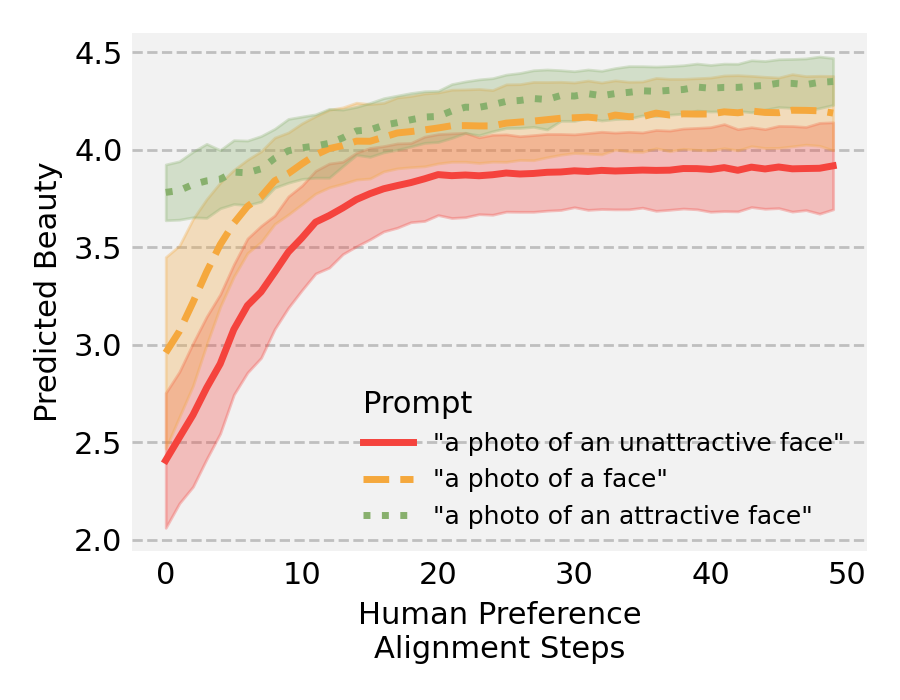}
  \end{subfigure}

  \caption{Human preference guided generation using different face-based prompts. Human preference alignment is done using ReNO with HPSv2.1 as the reward model and PixArt-$\alpha$ as the text-to-image model. The plot shows the average predicted facial attractiveness over 50 optimization steps.}
  \label{fig:reno_results_plot}
\end{figure}

\section{Additional Experimental Details}

\subsection{Facial Attractiveness Predictor}

In the main body of text we use a checkpoint finetuned from DINOv3 \citep{simeoni2025dinov3} to predict facial attractiveness. Specifically, we use the dinov3-vits16-pretrain-lvd1689m checkpoint as our starting point. We then finetune the checkpoint on cropped and aligned facial images from the CelebA \citep{liu2015faceattributes} and SCUT-FBP5500 \citep{liang2017SCUT} datasets. Each dataset has its own prediction head, with the binary labels from CelebA learned using binary cross entropy loss and the continuous labels from SCUT-FBP5500 learned using mean squared error loss. Both label sets are direct attractiveness ratings of \emph{real} photographs, collected independently of any AI-generated image or preference comparison task. We train for 20 epochs using the AdamW optimizer \citep{kingma2014adam} in PyTorch with a learning rate of 1e-5 for the backbone and a learning rate of 1e-3 for the prediction heads. \revision{Other optimizer parameters are set to PyTorch defaults: beta values are (0.9, 0.999) and  weight decay is 0.01.} The final checkpoint is taken from the epoch with the lowest total validation loss (90/10 training/validation split is used for both datasets). \revision{Figure~\ref{fig:beauty_loss_plots} shows the training and validation losses over the 20 epochs. The model converges quickly, reaching its peak validation loss after the second epoch. On the CelebA validation split, the model reaches 81\% accuracy on binary attractiveness prediction.}

\begin{figure*}[t]
  \centering

  % --- SUBFIGURE A: Preference-guided Generation Diagram ---
    \includegraphics[width=0.66\linewidth]{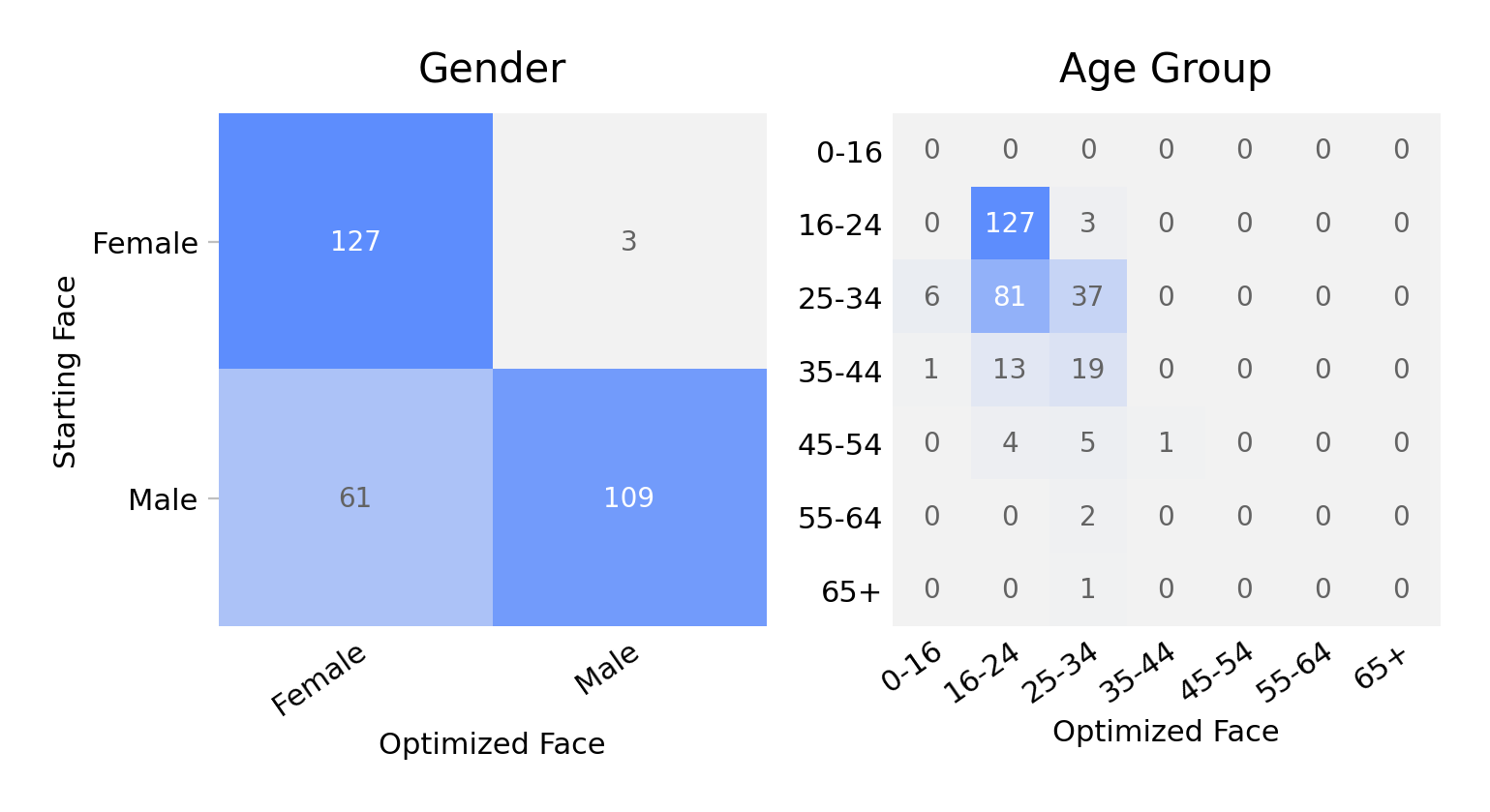}

  \caption{Confusion matrices that show the age and gender of faces before and after ReNO optimization. PixArt-$\alpha$ is the base text-to-image model used and HPSv2.1 is the reward model used.}
  \label{fig:confusion_reno}
\end{figure*}

\begin{figure*}[t]
  \centering
  \includegraphics[width=1.0\linewidth]{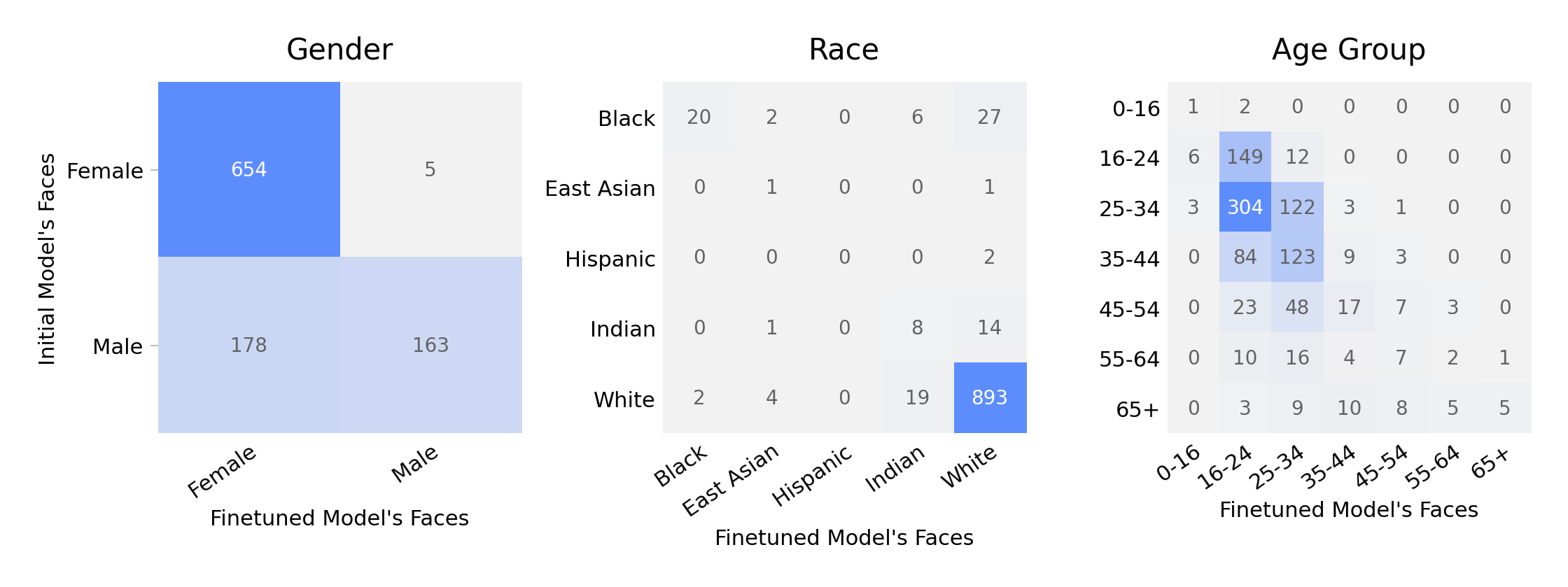}
  \caption{Confusion matrices that show the age, race, gender of faces before and after DRaFT finetuning. A DDPM model trained on CelebA faces is used as the base diffusion model and PickScore is used for reward model finetuning.}
  \label{fig:confusion_draft}
\end{figure*}

\begin{figure}[t]
  \centering
  \setlength{\tabcolsep}{1.5pt}
  \renewcommand{\arraystretch}{1}
  \begin{tabular}{@{}c@{\hspace{2pt}}c@{\hspace{2pt}}c@{\hspace{2pt}}c@{}}
    {\sffamily\footnotesize\bfseries\color{arrowgray} Epoch 0} &
    {\sffamily\footnotesize\bfseries\color{arrowgray} Epoch 1} &
    {\sffamily\footnotesize\bfseries\color{arrowgray} Epoch 2} &
    {\sffamily\footnotesize\bfseries\color{arrowgray} Epoch 3} \\[2pt]
    \includegraphics[width=0.235\columnwidth]{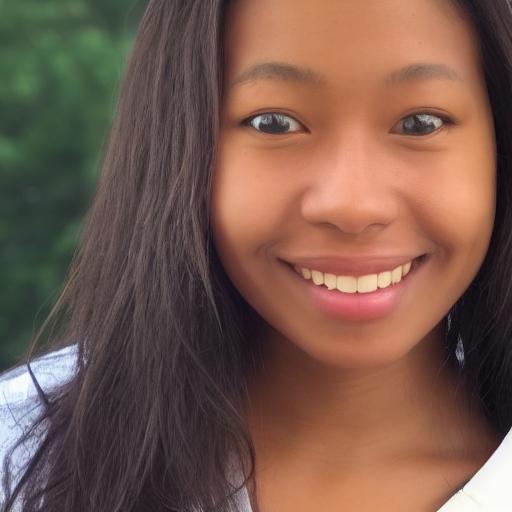} &
    \includegraphics[width=0.235\columnwidth]{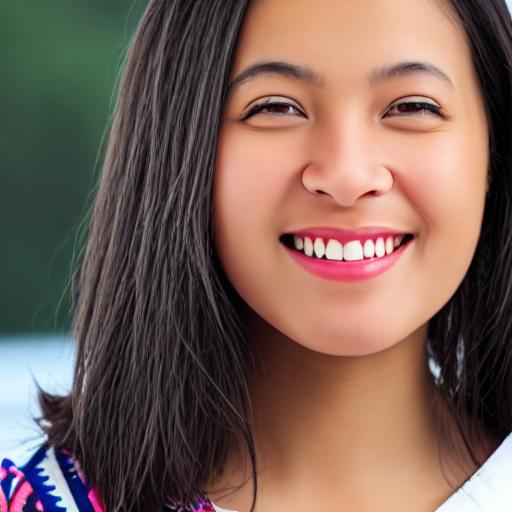} &
    \includegraphics[width=0.235\columnwidth]{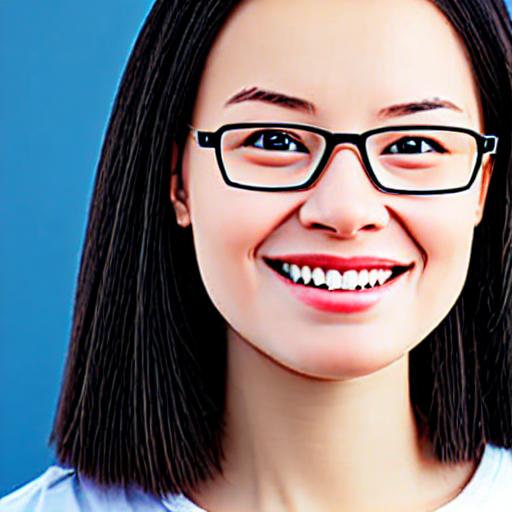} &
    \includegraphics[width=0.235\columnwidth]{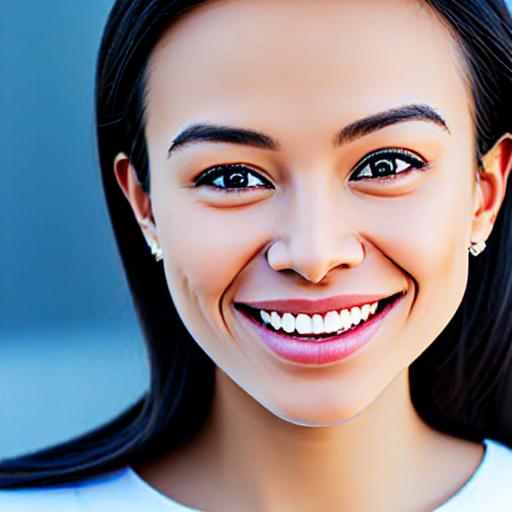} \\[-1.5pt]
    \includegraphics[width=0.235\columnwidth]{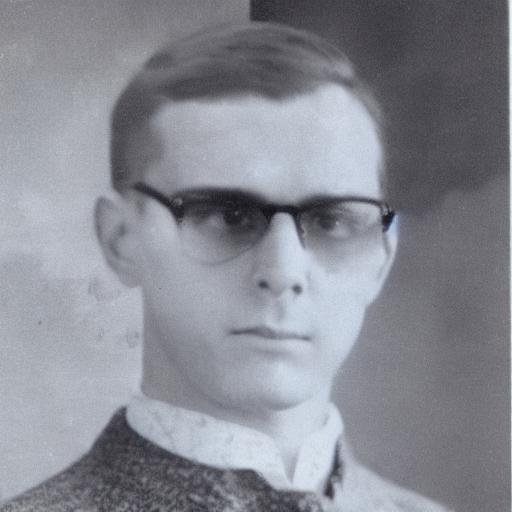} &
    \includegraphics[width=0.235\columnwidth]{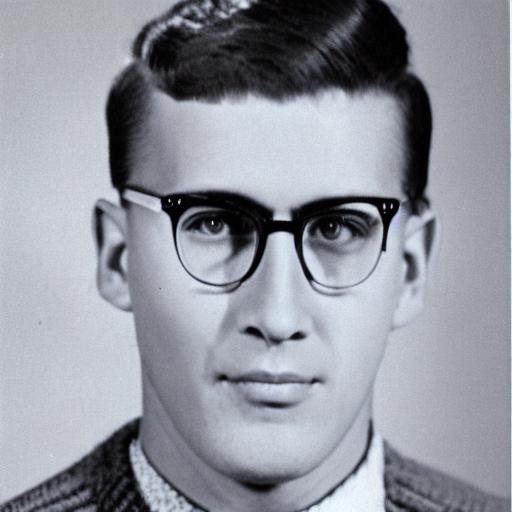} &
    \includegraphics[width=0.235\columnwidth]{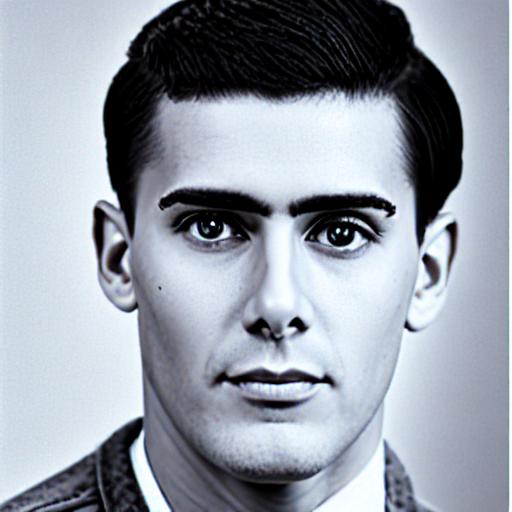} &
    \includegraphics[width=0.235\columnwidth]{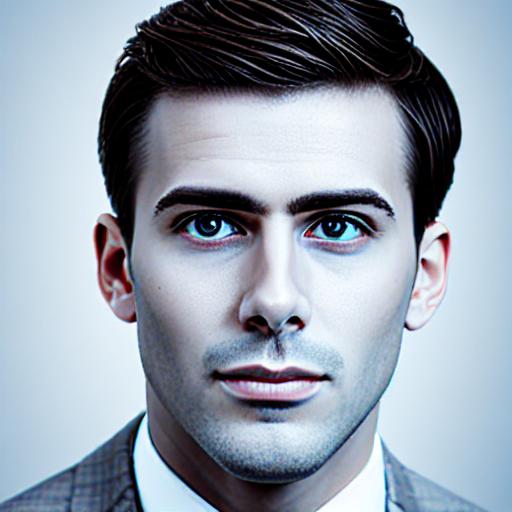} \\
  \end{tabular}
  \caption{Qualitative AlignProp finetuning results across epochs for two samples. The prompts used for generation are ``She has a extremely mild smile. This female is a young adult and has no eyeglasses. This female shows the whole forehead without any bangs.'' and ``There is not any glasses on his face. This person looks extremely young and has no mustache, and no bangs. This guy looks serious with no smile in his face.'' for the top and bottom rows, respectively.}
  \label{fig:finetuning_alignprop}
\end{figure}

\begin{figure}[t]
  \centering
  \setlength{\tabcolsep}{1.5pt}
  \renewcommand{\arraystretch}{1}
  \begin{tabular}{@{}c@{\hspace{2pt}}c@{\hspace{2pt}}c@{\hspace{2pt}}c@{}}
    {\sffamily\footnotesize\bfseries\color{arrowgray} Epoch 0} &
    {\sffamily\footnotesize\bfseries\color{arrowgray} Epoch 1} &
    {\sffamily\footnotesize\bfseries\color{arrowgray} Epoch 2} &
    {\sffamily\footnotesize\bfseries\color{arrowgray} Epoch 3} \\[2pt]
    \includegraphics[width=0.235\columnwidth]{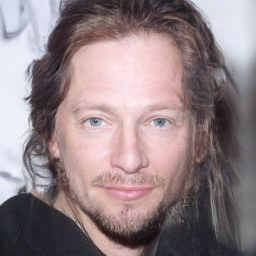} &
    \includegraphics[width=0.235\columnwidth]{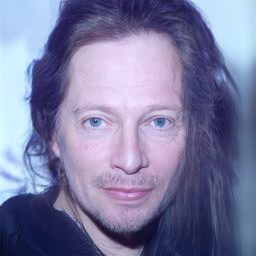} &
    \includegraphics[width=0.235\columnwidth]{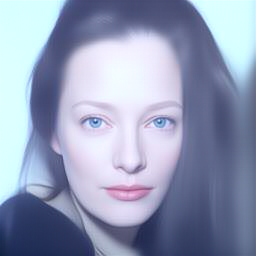} &
    \includegraphics[width=0.235\columnwidth]{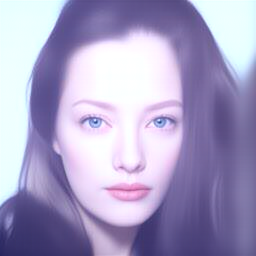} \\[-1.5pt]
    \includegraphics[width=0.235\columnwidth]{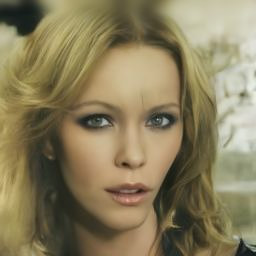} &
    \includegraphics[width=0.235\columnwidth]{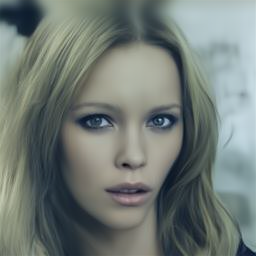} &
    \includegraphics[width=0.235\columnwidth]{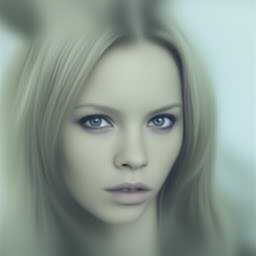} &
    \includegraphics[width=0.235\columnwidth]{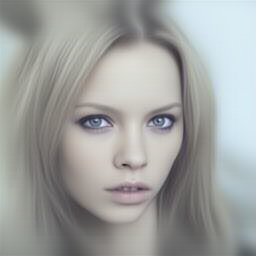} \\
  \end{tabular}
  \caption{Qualitative DRaFT finetuning results across epochs for two samples.}
  \label{fig:finetuning_draft}
\end{figure}

\subsection{ReNO Experiments}

In this section we provide further analysis on ReNO-based \citep{eyring2024reno} guided generation using HPSv2.1 \citep{wu2023human}. As in the main body of the text we use PixArt-$\alpha$ \citep{chen2024pixart} as the base text-to-image model. Specifically, we run ReNO for 50 steps using 3 distinct prompts: "a photo of an unattractive face", "a photo of a face", and "a photo of an attractive face". We apply ReNO to 100 images for each prompt. In Figure~\ref{fig:reno_results_plot} we provide the mean predicted beauty (SCUT-FBP5500 prediction head used) for images across all 50 steps. We can see that preference alignment boosts the attractiveness of all images regardless of the prompt used. We again emphasize how this could limit user control because whatever prompt is used, the faces generated will almost entirely appear as very attractive. Through preference alignment, the ability to generate unattractive faces is effectively removed.

We also provide analysis on how the optimization of these 300 images alters the age, race, and gender of the faces depicted. To measure this we use a pretrained classifier \footnote{\url{https://huggingface.co/TimmaJ/age-gender-race-prediction}} trained on FairFace \citep{karkkainen2021fairface} to predict the age, race, and gender of face before and after preference alignment. 

According to our classifier, all faces generated before and after generation were white. Figure~\ref{fig:confusion_reno} shows confusion matrices describing how both age and gender change during optimization. The majority of faces remain the same gender as when they started, however, out of the faces that do change, 61 change from male to female and only 3 change from female to male. Similarly, most faces stay in the same age category before and after optimization, but nearly all faces that changes age are changed to look younger, with an especially large focus on the 16-24 years old category. This suggests that HPSv2.1 shows a preference for young female faces.

\revision{All ReNO experiments use the default hyperparameters provided in their official GitHub repo unless stated otherwise.}

\subsection{Preference Finetuning Experiments}

In the main body of the text we also use DRaFT \citep{clark2024directly} to finetune a CelebA DDPM model \citep{ho2020denoising} using rewards from PickScore \citep{kirstain2023pick}. We measure how finetuning changes the demographics of the face using a  FairFace trained classifier. We generate 1000 images with the same random seeds, using the model weights pre- and post-finetuning. The confusion matrices in Figure~\ref{fig:confusion_draft} show how finetuning with PickScore impacts age, gender, and race. Similarly to Figure~\ref{fig:confusion_reno}, we note that faces tend to be feminized and made younger. The base models shows more diversity in terms of race, but it is common for non-white to be changed into white faces.

We also show qualitative results from our AlignProp and DRaFT finetuning experiment in Figures~\ref{fig:finetuning_alignprop} and \ref{fig:finetuning_draft}, respectively. We can see that both finetuning approaches result in a combination of different effects including the reduction of artifacts, the changing of demographics, and an increase in facial attractiveness.

\subsection{GPT-Image-2 Stable Diffusion Comparison}

Figure 1 in the main body of the paper includes two images, one generated by Stable Diffusion v1.5 and one generated by GPT-Image-2. The first image was generated using Stable Diffusion v1.5 using the prompt ``a photo of an attractive face''. To maintain the structure of this image, but change the facial attractiveness, we then blurred this image and gave it to GPT-Image-2 with the following prompt: ``Unblur this image according to the following prompt. ``a photo of an unattractive face'' ''. We then evaluated both images using reward models which all received a neutral prompt ``a photo of a face''. 

\revision{For this illustrative example we make the assumption that the image produced by GPT-Image-2 is of better quality than the image produced by Stable Diffusion v1.5. We believe this assumption is well founded due to GPT-Image-2 being significantly more modern than Stable Diffusion v1.5. In addition, visual inspection of the Stable Diffusion image reveals clear artifacts  such as misshapen irises in the eyes, unnatural looking teeth, and overly smoothed skin. In comparison, the image produced by GPT-Image-2 contains no obvious artifacts.}

\subsection{Additional User Study Details}
\label{sec:appendix_user_study_details}

\begin{figure}[t]
  \centering
  \begin{subfigure}{\columnwidth}
    \centering
    \includegraphics[width=0.9\linewidth]{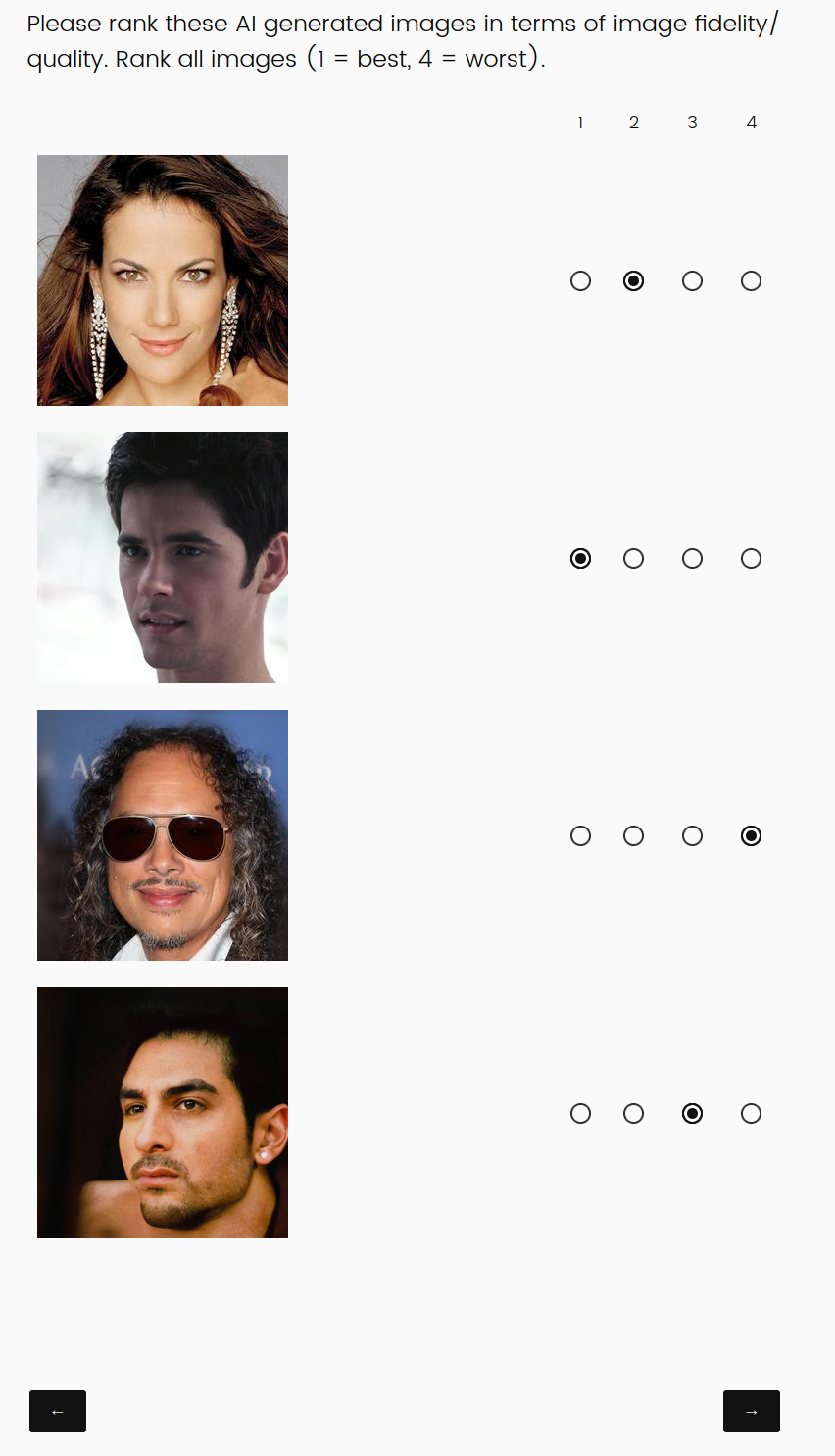}
  \end{subfigure}
  \caption{Screenshot from the first question of our Qualtrics user study.}
  \label{fig:screenshot_1}
\end{figure}

\begin{figure}[t]
  \centering
  \begin{subfigure}{\columnwidth}
    \centering
    \includegraphics[width=0.9\linewidth]{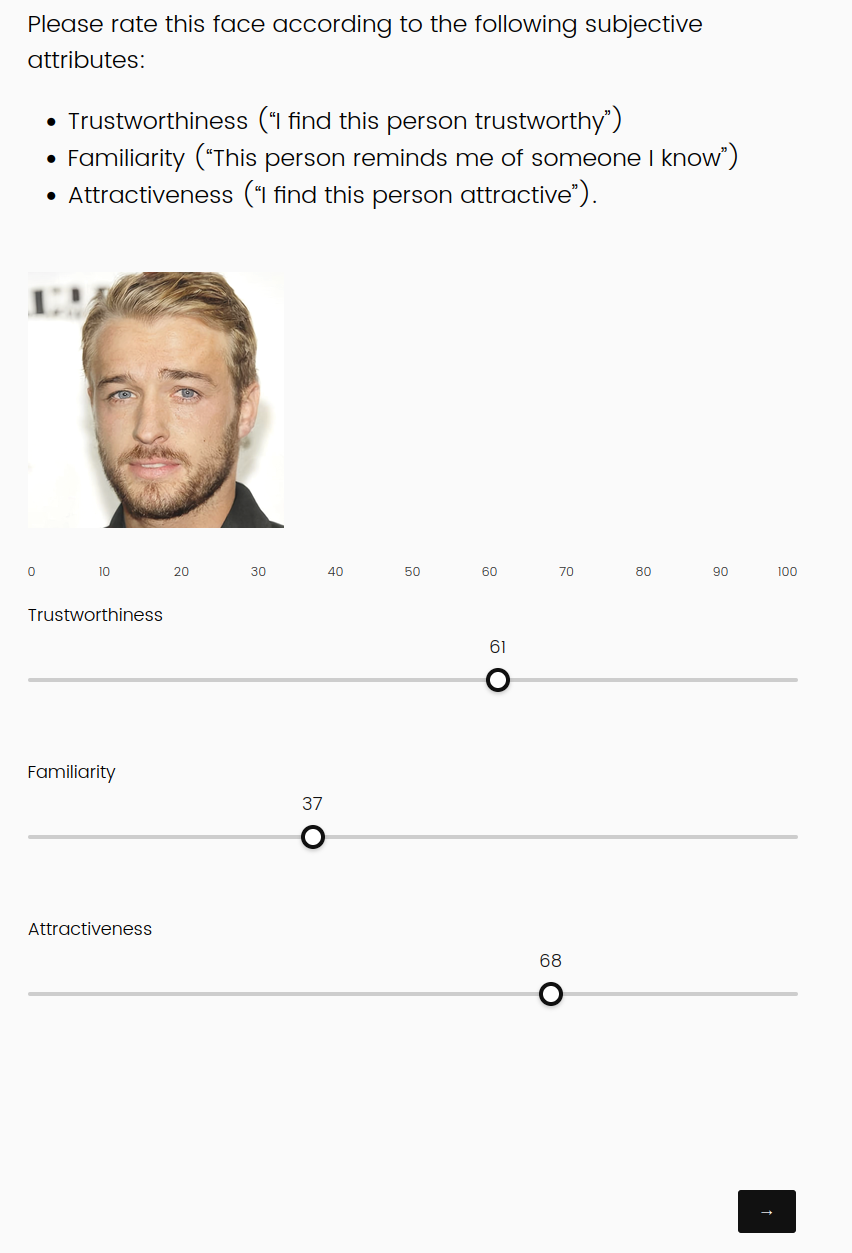}
  \end{subfigure}
  \caption{Screenshot from the second question of our Qualtrics user study.}
  \label{fig:screenshot_2}
\end{figure}

Our user  study was approved by the Institutional Review Board (IRB) at our institution. In accordance with the approved guidelines, all participants provided informed consent before beginning the online survey, and data collection was conducted completely anonymously with no personally identifiable information recorded. Participants were informed of their right to withdraw from the study at any point prior to submission. All participants were recruited through the Prolific platform. 

The use of deception was reviewed and approved by our IRB as necessary to the study design, and all participants were fully debriefed on submission, at which point they were informed that half of the images shown were real photographs from the CelebA dataset and given the opportunity to withdraw their responses.

Prior to completing the survey, users are prompted to read a consent form and click that they consented to taking part in the study. This includes a brief introduction to the task ``I am inviting you to participate in a study regarding how humans perceive, evaluate, and rate digital images of human faces. Specifically, This study aims to understand how different types of questioning and presentation formats affect how people judge the quality and characteristics of facial imagery. We are investigating the relationship between human visual preferences and image evaluation metrics.''.

In Figures~\ref{fig:screenshot_1} and \ref{fig:screenshot_2} we include screenshots of the first and second questions in our Qualtrics user study.

\revision{Our user study assumes that real images have comparable fidelity and that they are inherently more realistic than generated images. Here, we provide additional discussion and analysis to support this assumption.}

\revision{Firstly, blurring real images could confound the results of our user study. However, CelebA-HQ images are annotated with a binary ``Blurry'' attribute, and only 0.38\% of images in the dataset are labeled as ``Blurry''. Importantly, none of the real images used in our user study are labeled as ``Blurry''. This makes variations in image fidelity less likely for our real user study images.}

\begin{figure}[t]
\centering
\includegraphics[width=0.85\linewidth]{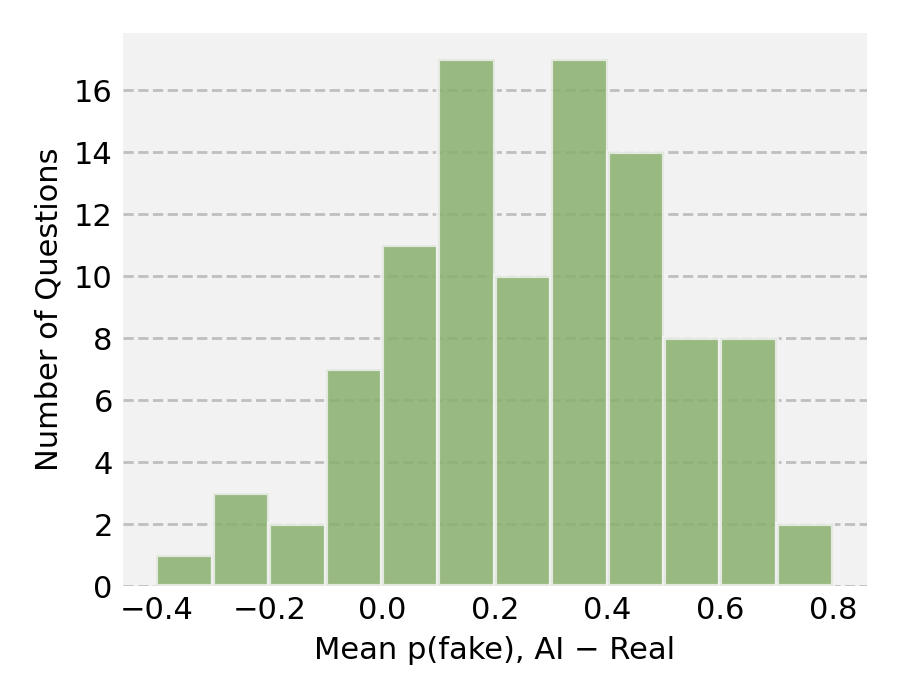}
\caption{\revision{Distribution of differences in MaskCLIP's estimated probability of an image being fake between AI-generated and real images across the 100 user-study questions. For each question, we calculate the mean $p(\text{fake})$ for AI-generated images and subtract the corresponding mean for real images. Positive values indicate that MaskCLIP assigns a higher probability of being fake to AI-generated images, while negative values indicate the opposite. The difference is positive for 87 of the 100 questions, suggesting that the detector generally distinguishes the real images as more likely to be real than the AI-generated images.}}
\label{fig}
\end{figure}

\revision{Secondly, although quantifying relative image quality between individual images is challenging, recent work on general-purpose AI-image detection, MaskCLIP \citep{wang2025opensdi}, provides a means of estimating whether an image is real or AI-generated. We therefore use MaskCLIP to estimate the probability that each image is AI-generated. We use these estimates as a rough proxy for perceived ``realness'', providing additional evidence relevant to our assumption. Specifically, we use the \texttt{MaskCLIP\_sd15\_20241109\_08\_53\_19} variant of MaskCLIP to obtain the ``real'' and ``fake'' logits, which are then passed through a softmax function to estimate the probability that an image is fake, $p(\text{fake})$. We calculate the mean $p(\text{fake})$ for real and AI-generated images separately within each of the 100 questions, and plot the difference between these means ($p(\text{fake})$ for AI images minus $p(\text{fake})$ for real images) in Figure~\ref{fig}. For 87 of the 100 questions, the difference is positive, indicating that MaskCLIP assigns a higher probability of being fake to the AI-generated images than to the real images. This provides further evidence that the real images are generally perceived as more realistic by the detector, although the estimates should not be interpreted as direct measurements of human-perceived realism.}

\revision{\subsection{Additional Leaderboard Simulations}}

% \revision{In this section, we provide results for the inverse of the Elo score simulation experiment in the main body of the paper. Specifically, we define two teams, the first containing more attractive real images and the second containing less attractive AI-generated images. As in the main body of the paper, images and attractiveness ratings are taken from the user study results. This configuration has a smaller gap in average attractiveness (63.5 for real images, 53.7 for AI-generated images), but importantly, real images are still ranked as more attractive than the AI images on average. We run 10,000 Elo ranking simulations with initial Elo scores of 1500 and a K-factor of 64.}

% \revision{Table~\ref{tab:elo_reversed} shows the results of this experiment. Overall, for all responses, this changes the winner to real images. This provides further supporting evidence of our claim that attractive faces can bias leaderboards.}

\begin{figure}[t]
  \centering
  % --- SUBFIGURE A: Preference-guided Generation Diagram ---
  \includegraphics[width=0.85\linewidth]{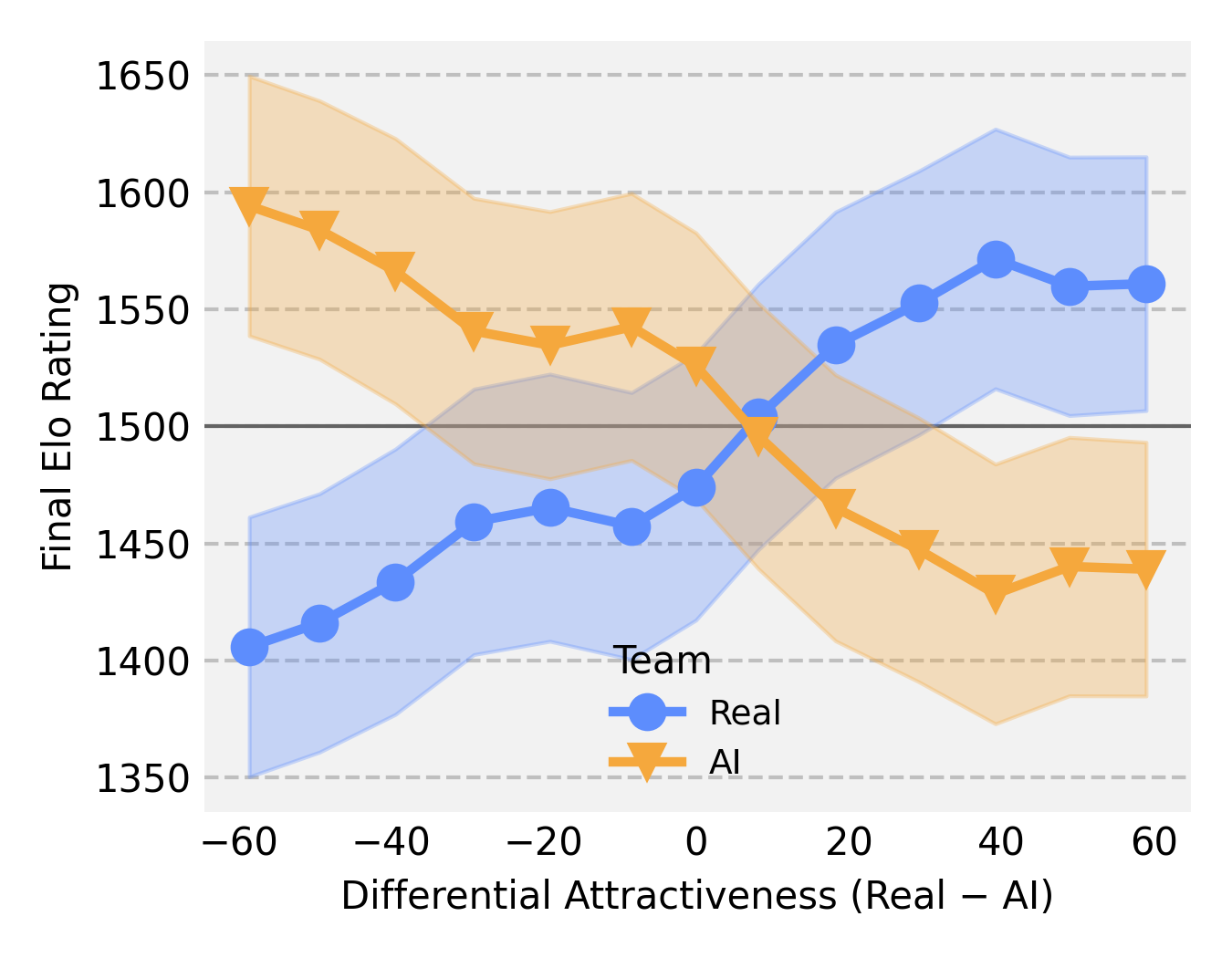}
  \caption{Average final Elo ratings against differential attractiveness (Real $-$ AI). Shaded regions indicate standard deviation within 10000 simulations. As real images become more attractive relative to AI-generated images, their average final Elo ratings increase, suggesting that attractiveness can systematically bias leaderboard rankings.}
  \label{fig:diff_attr_elo}
\end{figure}

\revision{In the main text, we use Elo simulations built from our user study preferences to examine how attractiveness affects leaderboard rankings. The results suggest that when two teams are compared, Elo ratings skew towards whichever team contains the more attractive faces. To investigate this relationship further, we filter the image pairings from our user study according to the difference in average attractiveness between the real and AI-generated images, allowing us to examine how Elo ratings vary across a range of relative attractiveness levels. For each level of differential attractiveness, we rerun the 10,000 Elo simulations using the corresponding filtered pairings. Figure~\ref{fig:diff_attr_elo} presents the resulting average final Elo ratings as a function of differential attractiveness. Although the results exhibit some variability due to the reduced number of available pairings at certain attractiveness levels, the overall trend indicates that teams with more attractive faces tend to achieve higher Elo ratings.}

% \begin{table}[t]
%   \footnotesize
%   \centering
%   \setlength{\tabcolsep}{8pt}
%   \renewcommand{\arraystretch}{1.25}
%   \begin{tabular}{@{}lcc@{}}
%     \toprule
%     \textbf{Instruction Type} & \textbf{Real Images} & \textbf{AI Images} \\
%     \midrule
%     Fidelity Instruction   & $1498.6 \pm 0.6$ & $\mathbf{1501.4} \pm 0.6$ \\
%     Aesthetic Instruction & $\mathbf{1512.1} \pm 0.6$ & $1487.9 \pm 0.6$ \\
%     All Responses   & $\mathbf{1505.8} \pm 0.6$ & $1494.2 \pm 0.6$ \\
%     \bottomrule
%   \end{tabular}
%   \caption{\revision{Mean Elo ratings \revision{(± standard error)} when comparing less attractive AI-generated CelebA-HQ images against more attractive real CelebA-HQ images. User study preference data is used to simulate randomized matchmaking between the two groups over 10000 runs. We report the simulated outcomes across ``fidelity'' instructions, ``aesthetic'' instructions, and combined responses.}}
%   \label{tab:elo_reversed}
% \end{table}

\revision{\subsection{Additional Mitigation Details}}

\revision{The generated images, along with their corresponding prompts and unadjusted scores can be downloaded from: \url{https://huggingface.co/datasets/harveymannering/t2i-faces}.}

\revision{\section{Open Source Leaderboards}}

\revision{In the main body of the text, we make the claim that the majority of models on LMArena's T2I open-weight leaderboard have had some form of human preference alignment during post-training. To substantiate this claim, we enumerate all 18 of the models and their leaderboard placements at the time of writing (August 2026) below:
\begin{enumerate}
    \item ideogram-4.0-quality \citep{ideogram-4-2026}
    \item Cosmos3-Super-Text2Image (Agentic) \citep{agarwal2026cosmos}
    \item Cosmos3-Super-Text2Image
    \item hunyuan-image-3.0 \citep{cao2025hunyuanimage}
    \item flux-2-dev \citep{flux-2-2025}
    \item qwen-image-2512 \citep{wu2025qwen}
    \item hidream-o1-image \citep{hidreamolimage}
    \item krea-2-turbo \citep{krea2026technicalreport}
    \item z-image-turbo \citep{team2025zimage}
    \item flux-2-klein-9b
    \item qwen-image-prompt-extend
    \item qwen-image
    \item flux-2-klein-4b
    \item glm-image \citep{glmimage}
    \item flux-1-dev-fp8 \citep{flux2024} 
    \item flux-1-kontext-dev \citep{labs2025flux}
    \item stable-diffusion-v35-large \citep{esser2024scaling}
    \item bagel \citep{deng2025emerging}
\end{enumerate}}

\revision{We consider the technical reports and blog posts accompanying each of these models' releases. Models that have undergone some sort of preference alignment include: Cosmos3-Super-Text2Image (Agentic), Cosmos3-Super-Text2Image, hunyuan-image-3.0, qwen-image-2512, hidream-o1-image, krea-2-turbo, z-image-turbo, qwen-image-prompt-extend, qwen-image, and glm-image. Models that do not perform preference alignment include: flux-1-kontext-dev, stable-diffusion-v35-large, and bagel. Models for which we could not discern from the documentation if any preference alignment had taken place include: ideogram-4.0-quality, flux-2-dev, flux-2-klein-9b, flux-2-klein-4b, and flux-1-dev-fp8.}

\revision{So, out of the 18 models, 10 have undergone preference alignment, 3 have not, and 5 are indeterminate. Therefore, the majority of models have had some form of preference alignment post-training.}

% {
%     \small
%     \bibliographystyle{ieeenat_fullname}
%     \bibliography{main}
% }

% \end{document}